\documentclass{article}

\usepackage{microtype}
\usepackage{subcaption,graphicx}
\usepackage{booktabs} 
\usepackage[a4paper]{geometry}

\usepackage{hyperref}
\hypersetup{
    colorlinks=true,
    linkcolor=blue,
    citecolor = blue
}

\usepackage{natbib}
\usepackage{amsmath}
\usepackage{amssymb}
\usepackage{mathtools}
\usepackage{amsthm}
\usepackage{tikz}
\usetikzlibrary{cd}
\usepackage{adjustbox}

\usepackage{authblk}
\usepackage{float}

\usepackage[capitalize,noabbrev]{cleveref}

\theoremstyle{plain}

\theoremstyle{definition}

\theoremstyle{remark}

\usepackage[textsize=tiny]{todonotes}

\title{Bipartite Graph Variational Auto-Encoder with Fair Latent Representation
to Account for Sampling Bias in Ecological Networks}
\author[1]{Emre Anakok}
\author[2]{Pierre Barbillon}
\author[3]{Colin Fontaine}
\author[4]{Elisa Thebault}
\affil[1,2]{Université Paris-Saclay, AgroParisTech, INRAE, UMR MIA Paris-Saclay, 91120, Palaiseau, France.}
\affil[3]{Centre d'Écologie et des Sciences de la Conservation, MNHN, CNRS, SU, 43 rue Buffon, 75005 Paris, France}
\affil[4]{Sorbonne Université, CNRS, IRD, INRAE, Université Paris Est Créteil, Université Paris Cité, Institute of Ecology and Environmental Sciences (iEES-Paris), 75005 Paris, France}
\affil[1]{emre.anakok@agroparistech.fr}

\date{}                     
\begin{document}
\maketitle

\begin{abstract} 

Citizen science monitoring programs can generate large amounts of valuable data, but are often affected by sampling bias. We focus on a citizen science initiative that records plant-pollinator interactions, with the goal of learning embeddings that summarize the observed interactions while accounting for such bias.
In our approach, plant and pollinator species are embedded based on their probability of interaction. These embeddings are derived using an adaptation of variational graph autoencoders for bipartite graphs. To mitigate the influence of sampling bias, we incorporate the Hilbert-Schmidt Independence Criterion (HSIC) to ensure independence from continuous variables related to the sampling process. This allows us to integrate a fairness perspective—commonly explored in the social sciences—into the analysis of ecological data.
We validate our method through a simulation study replicating key aspects of the sampling process and demonstrate its applicability and effectiveness using the Spipoll dataset.

\end{abstract}

\section{Introduction}

Citizen science programs facilitate the accumulation of biological or ecological data and are progressively more prevalent in biodiversity monitoring efforts \citep{conradReviewCitizenScience2011,chandlerContributionCitizenScience2017,pocockVisionGlobalBiodiversity2018a}. However, these methods of collecting data are prone to sampling bias due to the multiplication of observers and associated observer effects \citep{birdStatisticalSolutionsError2014}. Further, one notable bias encountered in citizen science programs is  the progressive accumulation of user experience in data collection \citep{kellingCanObservationSkills2015}. For example, \citet{jiguetMethodLearningCaused2009} found in a breeding bird survey that observers tended to count more birds after the first year compared to the initial year of observation. In order to mitigate the bias introduced by the observers, \citet{johnstonEstimatesObserverExpertise2018} suggested estimating the observers' expertise and incorporate it as a covariate in their model.

As a study case, we focus on a citizen science based monitoring schemes, named Spipoll, where participants collect data on plant pollinator interactions across France. These data allow an unprecedented understanding of the variations of plant-pollinator ecological networks \citep{deguinesWhereaboutsFlowerVisitors2012,pocockVisualisation2016}. 

As other dataset collected using citizen science, the Spipoll dataset is prone to bias caused by variations in users skills and experience.

Given the amount of observed interactions and the need for bias mitigation, network analysis on this very large plant-pollinator network  could profit from recent developments in machine learning such as graph embeddings using graph neural networks (GNN) and a dedicated bias mitigation strategy.

Here, we propose a framework to reconstruct plant-pollinator network from such citizen science biodiversity monitoring schemes while accounting for associated sampling bias.

\paragraph{Presentation of the Spipoll data set and its sampling biases}
Spipoll is a French citizen science program aiming at monitoring plant-pollinator interactions across metropolitan France since 2010. This monitoring follows a simple protocol. Briefly, volunteers can choose a flowering plant where and when they like, and during 20 minutes, take pictures of all different insects that land on the flowers of the monitored plant. Then, using an online identification tool, they identify each different insect that has been photographed and upload their data on a dedicated website. Each session is thus a set of insect interactions with a given plant species that have been observed at a given time and place, and by a given volunteer whose specific skills could affect the quality of the observation.

\begin{figure}
    \centering
    \includegraphics[width=0.5\textwidth]{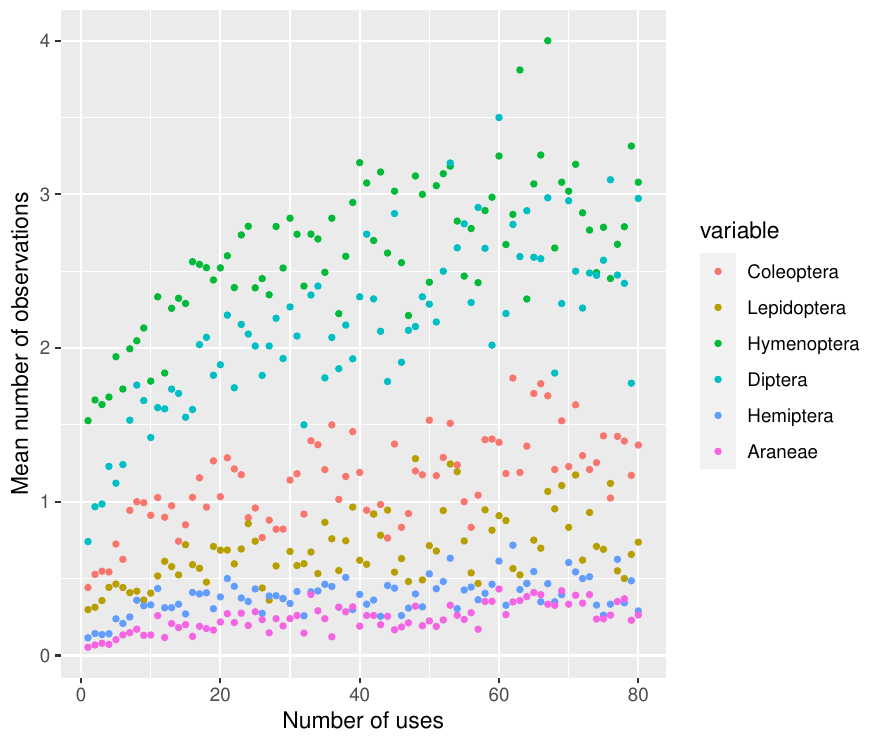}
    \caption{Average number of observations per insect order as a function of observer experience level in the Spipoll data set. 
    }
    \label{FIG0}
\end{figure}

One noticeable bias in the Spipoll dataset is that observers tend to record more pollinating insects per observation session as they get more experienced. This is illustrated in \cref{FIG0} where the average number of observed insects per session increases with the number of time the user has performed the protocol, with different slopes depending on the insect order. Such bias is related to both the ability of observers to differentiate and identify the various insect species which has been shown to increase with observer experience \citet{deguinesFosteringCloseEncounters2018} and the increasing photographing skills that allow the observers not to miss small and/or fast moving pollinators. In this context, \citet{deguinesFunctionalHomogenizationFlower2016, deguinesFosteringCloseEncounters2018} addressed the potential variation in the number of photographed insects among  
different observers by incorporating observer identity as random terms in their models.

\paragraph{Graph embedding}
Graph embedding regroups different methods, allowing to represent a network into a vector space in order to gain understanding of key network features. These methods are especially important in the context of large networks.  
Recently developed graph neural networks (GNNs) enable graph embedding with large-scale methods such as graph isomorphism network \citep{xuHowPowerfulAre2019}, graph attention network \citep{velickovicGraphAttentionNetworks2018} or the variational graph auto-encoder \citep{Kipf}. All these methods can also handle numerous covariates on nodes.

GNNs are currently growing in popularity in various domains such as bioinformatics \citep{zhangGraphNeuralNetworks2021} and chemistry \citep{reiserGraphNeuralNetworks2022}. 
In ecology, networks have been analyzed to study various types of interaction between species of plants and animals \citep{ings_review_2009}. The stochastic block model \citep{nowickiEstimationPredictionStochastic2001} and the latent block model \citep{govaert2010latent} for bipartite graphs are notorious models using latent variable in ecology \citep{terry_finding_2020,durand-bessart_trait_2023}. 
While graph embedding methods start being used for ecological networks (e.g. \citealt{botellaAppraisalGraphEmbeddings2022}, \citealt{strydomFoodWebReconstruction2022}), GNNs have yet to be diffused in that research field. GNNs could be particularly relevant for ecological networks because very large data sets of interactions among species are now becoming available (e.g. through the development of citizen science programs), in addition to many covariates linked to the nodes (e.g. species name and traits, environmental characteristics at the time of interaction observation).
An important issue with the analysis of ecological networks is related with the strong effects of sampling effort and methods on network structure \citep{jordano_sampling_2016,dore_relative_2021}. 
Sampling interactions among species to build ecological networks is indeed a daunting task. For example, extensive sampling of plant-pollinator interactions has been shown to only provide a subset of the existing interactions \citep{chacoff_evaluating_2012,jordano_sampling_2016} and sampling protocols are known to be subject to bias \citep{GibsonSampling2011}, both of which need to be accounted for to make meaningful ecological interpretations \citep{dore_relative_2021}.   
One could wish to have an embedding which is independent of a certain set of covariates linked to such sampling effects and related bias. This can be of particular interest for citizen science programs, where biases can arise from the observer's experience level \citep{jiguetMethodLearningCaused2009,deguinesFosteringCloseEncounters2018}.

\paragraph{Related works for correcting bias}
The solution to correct for sampling biases in ecological network analysis may be found in machine learning applied to sociology with the notion of fairness \citep{catonFairnessMachineLearning2020,careyFairnessFieldGuide2022}, where the main idea is to train an algorithm whose predictions are independent of a protected variable. It is used in sociology \citep{careyFairnessFieldGuide2022} to have predictions that are independent of gender, sexuality or disability. Recently, works on fairness have been extended to social network analysis \citep{saxenaFairSNAAlgorithmicFairness2022}, for fair link prediction \citep{liDYADICFAIRNESSEXPLORING2021}, fair graph exploration \citep{rahmanFairwalkFairGraph2019}.
Fairness has also been developed for variational auto-encoder, with the fair auto-encoder \citep{louizosVariationalFairAutoencoder2017} or with adversarial debiasing \citep{zhangMitigatingUnwantedBiases2018}, which can help to aim for a fair latent representation of the network.
Most papers about fairness for networks seek for fairness regarding a binary (or a categorical) protected variable \citep{saxenaFairSNAAlgorithmicFairness2022},  which is not always appropriate to the study of ecological networks where sampling biases might also be measured by continuous variables (e.g. sampling time, observer's experience level).
The adversarial debiasing from \citet{zhangMitigatingUnwantedBiases2018} could be applied to continuous variables, but has theoretical guarantees for the discrete protected variable. Even if the fair variational auto-encoder \citep{louizosVariationalFairAutoencoder2017} assumes independent prior, they penalize their loss with a regularization term that considers a discrete protected variable.  
In recommender systems \citep{beelResearchpaperRecommenderSystems2016, li2023recent}  where bipartite networks are also encountered,
different notions of fairness 
for different characteristics on graphs with heterogeneous nodes \citep{liFairnessRecommendationFoundations2023}      
are studied
but the protected variable is  binary or categorical.
However, the targeted fairness which is relevant in recommender system, is not adapted to address sample bias in ecology since the former focuses on ensuring equitable recommendations for users \citep{liFairnessRecommendationFoundations2023}, which is a user-centric goal, while the latter deals with a broader scope of the whole network.

There are other machine learning methods related with debiasing taking into account continuous variables, such as disentanglement methods. For example the $\beta-VAE$ \citep{Higgins2016betaVAELB} can provide a latent space with some conditionally independent latent factors. However their goal is different because they do not seek complete independence between the latent representation and the sensitive factors \citep{locatello2019fairness}. 
The method we propose relies on the Hilbert-Schmidt independence criterion (HSIC, \citealt{grettonMeasuringStatisticalDependence2005a}). The HSIC is a measure of independence between two random variables and can be used as a penalty term to ensure independence between continuous variable.
This metric has been put in practice in learning context by \citet{greenfeldRobustLearningHilbertSchmidt2020}, \citet{maHSICBottleneckDeep2020}, \citet{perez-suayFairKernelLearning2017} or \citet{quadriantoDiscoveringFairRepresentations2019}.

\paragraph{Outline of the paper}
In this paper, we aim to obtain a latent representation of the ecological network derived from Spipoll that is independent of covariates that may be continuous. 
After presenting the necessary background on GNNs and the HSIC, we introduce the model that makes recourse to an extension of the variational graph auto-encoder to bipartite networks with a penalization based on the HSIC. We attest the performance of our proposed model with simulated data mimicking the context of Spipoll. Finally, we apply our methodology to the Spipoll data set 
, where we show that accounting for the sampling effort can change the understanding of the network. 

\section{Bipartite and fair variational auto-encoder}


Our goal is to obtain an embedding of the nodes of the network. Since data collection is prone to sampling bias, our aim is to constrain the embedding to be independent of some variables related to the sampling condition.
These variables correspond to what are called protected variables in a fairness paradigm.
In the following, we present an extension of the variational graph auto-encoder (VGAE) \citep{Kipf} to a bipartite network. Then, since ensuring fairness in the network context is difficult because it involves addressing disparities in the representation and impact of the connection pattern, we rely on the optimization of a variational loss that incorporates a statistical measure of dependence between random variables, namely the HSIC \citep{grettonMeasuringStatisticalDependence2005a}. The HSIC and its efficient computation in high dimension are exposed in the following subsection. Finally, the optimized loss to ensure the embedding with the indepence constrain is presented in the final subsection.

\subsection{Bipartite variational graph auto-encoder}
We 
adapt the variational graph auto-encoder from \citet{Kipf} to the bipartite case by considering two graph convolutional networks (GCN), one for each node type. 
We consider a biadjacency matrix $B$ of size $n_1 \times n_2$ representing our bipartite graph. For all $i$ and $j$, $B_{i,j}\in \{0,1\}$ denotes the absence or the presence of interaction between the $i^{th}$ node of the first group and the $j^{th}$ node of the second group. Let 
$$D_1 = diag\left(\sum_{j=1}^{n_2}B_{i,j} \right),  \:  D_2 = diag\left(\sum_{i=1}^{n_1}B_{i,j} \right)$$  be respectively the row and the column degree matrices. We consider the normalized biadjacency matrix $\Tilde{B} = D_1^{-\frac{1}{2}} B D_2^{-\frac{1}{2}} $.
Additionally, $X_1$ is a $n_1 \times d_1$ matrix of node features for the first category, and $X_2$ is a $n_2 \times d_2$ matrix of node features for the second.

\subsubsection{Encoder}
\label{sec:encoder}

The encoder consists in associating latent variables for each node of both categories.
We denote by $Z_1$ a $n_1 \times D$ matrix, the rows of which $(Z_{1i}\in\mathbb{R}^D)_{1\le i\le n_1}$ are the latent variables associated to the nodes of the first category. Similarly, $Z_2$ is a $n_2 \times D$ matrix with rows  $(Z_{2j}\in\mathbb{R}^D)_{1\le j\le n_2}$ being the latent variables for nodes of the second category.

Our encoder is then defined as 
$$q(Z_1,Z_2|X_1,X_2,B) =  \prod_{i=1}^{n_1} q_{1}(Z_{1i}|X_1,B)\prod_{j=1}^{n_2} q_2(Z_{2j}|X_2,B) $$
where $q_1$ and $q_2$ correspond to multivariate normal distributions $\mathcal{N}(\mu,diag(\sigma^2))$.
 The parameters for the distributions $q_1$:
$(\mu_{1i},\log(\sigma_{1i}))_{1\le i\le n_1}\in \mathbb{R}^D\times \mathbb{R}^D$
are obtained by two GCN \citep{Kipf}, namely GCN$_{\mu_1}(X_1,B)$ and GCN$_{\sigma_1}(X_1,B)$
where:
$$\text{GCN}_{\mu_1}(X_1,B) = \Tilde{B}\text{ReLU}(\Tilde{B}^\top X_1 W^{(1)}_{\mu_1})W^{(2)}_{\mu_1}$$

with ReLU$(x) = max(x,0)$ and the weight matrices $W^{(k)}_{\mu_1}$ are to be estimated. GCN$_{\sigma_1}(X_1,B)$ is identically defined but with weight matrices $W^{(k)}_{\sigma_1}$. As \citep{Kipf}, we enforce that GCN$_{\mu_1}(X_1,B)$ and GCN$_{\sigma_1}(X_1,B)$ share the same first layer parameters, meaning that $W^{(1)}_{\mu_1}=W^{(1)}_{\sigma_1}$.
Symmetrically, the parameters for \\${q_2: (\mu_{2j},\log(\sigma_{2j}))_{1\le j\le n_2}\in \mathbb{R}^D\times \mathbb{R}^D}$ are obtained by two GCN, namely GCN$_{\mu_2}(X_2,B)$ and GCN$_{\sigma_2}(X_2,B)$ where
$$\text{GCN}_{\mu_2}(X_2,B) = \Tilde{B}^\top\text{ReLU}(\Tilde{B} X_2 W^{(1)}_{\mu_2})W^{(2)}_{\mu_2}.$$ 
GCN$_{\sigma_2}(X_2,B)$ is identical but with weight matrices $W^{(k)}_{\sigma_2}$, and with $W^{(1)}_{\mu_2}=W^{(1)}_{\sigma_2}$.

\subsubsection{Decoder}

Following \cite{rubin-delanchyStatisticalInterpretationSpectral2021}, we decide to use as a decoder the generalised random dot product

$$p(B|Z_1,Z_2) = \prod_{i=1}^{n_1}\prod_{j=1}^{n_2} p(B_{i,j}|Z_{1i},Z_{2j})$$
with 

$p(B_{i,j}|Z_{1i},Z_{2j}) = sigmoid( Z_{1i}^\top\mathbf{I}_{D_+,D_-}Z_{2j})$
where $sigmoid :x\mapsto \frac{1}{1+e^{-x}}$ and $\mathbf{I}_{D_+,D_-}$ is a diagonal matrix with $D_+$ ones followed by $D_-$ minus ones on its diagonal, such as $D_+ + D_-  = D $. We selected this decoder because its versatility covers a large family of graph structures.
The full auto-encoder can be summarized as

$$B,X_1,X_2\xrightarrow[encoder]{q(Z_1,Z_2|X_1,X_2,B)}Z_1,Z_2\xrightarrow[decoder]{p(B|Z_1,Z_2)}\widehat{B}.$$

\subsection{Hilbert Schmidt Independence Criterion}
\subsubsection{Definition}
Let $X$ and $Y$ two random variables in $\mathcal{X}$ and $\mathcal{Y}$ and $(X,Y)$ be the joint probability distribution. Let $\mathcal{F}$ and $\mathcal{G}$ be the RKHS on $\mathcal{X}$ and $\mathcal{Y}$ with their associated kernel $K : \mathcal{X} \times \mathcal{X} \to \mathbb{R}$ and $L : \mathcal{Y} \times \mathcal{Y} \to \mathbb{R}$. \citet{grettonMeasuringStatisticalDependence2005a} define the HSIC as the norm of the cross-variance operator between the distribution in the RKHS:
\begin{align}
    HSIC(X,Y) &= ||C_{X,Y} ||^2\notag\\
    &= \mathbb{E}_{XYX'Y'} [K(X,X')L(Y,Y')]
    + \mathbb{E}_{XX}[K(X,X')]\mathbb{E}_{YY'}[L(Y,Y')]\notag\\
    &- 2\mathbb{E}_{XY}[\mathbb{E}_{X'}[K(X,X')] \mathbb{E}_{Y'}[L(Y,Y')]].\notag
\end{align}
Using some specific kernels such as the Gaussian kernel $K(x_i,x_j) = e ^{-\frac{||x_i-x_j||^2}{2\sigma^2}}$ it can be shown that $HSIC(X,Y)= 0 \iff X\bot Y$\citep{grettonMeasuringStatisticalDependence2005a}.
\subsubsection{Estimation}

Given $(x_1,y_1)\dots(x_n,y_n)$ an i.i.d. sample drawn from $(X,Y)$, and given the corresponding evaluations of the two kernels $K_{ij} = K(x_i,x_j)$ and $L_{i,j} = L(y_i,y_j)$, a biased estimator of the HSIC is given by 
\begin{align}
&\widehat{HSIC}:= \widehat{HSIC}(\{(x_i,y_i)\}_{i=1}^n)  \notag\\
&=\frac{1}{n^2}\sum_{1\leq i,j\leq n}K_{i,j}L_{i,j}+\frac{1}{n^4}\sum_{1\leq i,j,p,q\leq n}K_{i,j}L_{p,q} - \frac{2}{n^3}\sum_{1\leq i,j,q\leq n}K_{i,j}L_{i,q}.
\end{align}
Under the assumption that $X$ and $Y$ are independent, it has been proved that the distribution of $n \times \widehat{HSIC}$ can be asymptotically approximated by a Gamma distribution \citep{gretton_gamma}:

$$n\times \widehat{HSIC} \sim \frac{x^{\alpha-1}e^{-\frac{x}{\beta}}}{\beta^\alpha \Gamma(\alpha)}
\text{\quad with } \alpha=\frac{\mathbb{E}[\widehat{HSIC}]^2}{\mathbb{V}[\widehat{HSIC}]}  \text{ and } \beta=\frac{n\mathbb{V}[\widehat{HSIC}] }{\mathbb{E}[\widehat{HSIC}]}. $$

During the learning phase, using the HSIC as a penalty between the coordinates in the latent space and the protected variable will assure that the latent space is as much as possible independent of the protected variable. At the end of the learning, we can check the independence by calculating the p-value of the test and compare it to the desired significance level. An example in the linear case is available in \ref{PCA with indep}.

\subsection{Estimation in high dimension}
 
Calculating $\widehat{HSIC}$ requires to compute the $n \times n$ Gram matrix and can be time-consuming for large value of $n$.
As a substitute, we can use Random Fourier Features (RFF) \citep{rahimiRandomFeaturesLargeScale2007}. RFF for learning has been used in \citep{louizosVariationalFairAutoencoder2017} to minimize the maximum mean discrepancy (MMD) between two distributions, in a setting where the protected variable have two possible values. Using the Gaussian kernel, minimizing the MMD is equivalent to matching all the moments of the two distribution, making the outcome of the encoder fair. In our settings, the HSIC can be seen as the MMD between the joint distribution $(X,Y)$ and the product of the distribution $X$ and $Y$. If the joint distribution is close to the product of the two distribution then the $X$ and $Y$ will behave as they are independent.

Assume that $K(x_i,x_j) = e^{-\frac{1}{2}||x_i-x_j||^2}$ ($\sigma^2=1$), with $x_i \in \mathbb{R}^d$, let $h < n$ be an integer, $\omega$ be a $h \times d$ matrix where all entries are independently drawn from $\mathcal{N}(0,1)$ and  $b$ be $d$-dimensional vector with each entry independently drawn  from $Unif([0,2\pi])$. For any $x_i$, its RFF is defined as 

$$\psi_X(x_i) = \sqrt{\frac{2}{h}}cos(\omega x_i^\top +b) \in \mathbb{R}^h.$$

The main idea behind this RFF representation is that $$K(x_i,x_j) \approx \langle \psi_X(x_i), \psi_X(x_j)\rangle.$$ This property will allow the computation to be much faster (see \ref{HSIC_temporal_gain}), while having a small error term \citep{sutherlandErrorRandomFourier2015}. If we also define $\psi_Y$ the as RFF of $Y$, then finally we can estimate the RFF HSIC \citep{zhangLargescaleKernelMethods2018} as done in \citet{quadriantoDiscoveringFairRepresentations2019}:

\begin{align}
    RFF \: HSIC = &\frac{1}{n^2}\left|\left|\sum_{i=1}^n\psi_X(x_i)\psi_Y(y_i)^\top  \right.\right.\left.\left.-\frac{1}{n}\left(\sum_{i=1}^n\psi_X(x_i) \right) \left(\sum_{i=1}^n\psi_Y(y_i) \right)^\top \right|\right|^2_2.
    \label{eq_RFF}
\end{align}

 The RFF HSIC can then be computed accurately with complexity $O(h^2)$. Several other methods of estimation for HSIC exists for large-scale problem \citep{zhangLargescaleKernelMethods2018}.
 For the rest of this paper, $h=100$ will provide enough accuracy.

 \subsection{Bipartite and fair graph variational auto-encoder}
\label{BIPARTITE AND FAIR GRAPH VARIATIONAL AUTO ENCODER}

Suppose our goal is to construct the latent representation $Z_1$ of the bipartite variational auto-encoder such as it is independent of a protected variable denoted by $S$. We optimize our parameters $(W_{v,i})_{1\leq v,i\leq 2}$ as defined in \cref{sec:encoder} to minimize a compromise between the variational lower bound of the auto-encoder and the HSIC between $\mu_1$ and the protected variable $S$. Even if the reconstruction would be penalized, this would yield a latent-space $\Tilde{Z_1} \sim \mathcal{N}(\mu_1,diag(\sigma^2_{1}))$ independent of $S$. 
 The complete loss of this auto-encoder can be written as:
\begin{align}
\label{loss1}
 L_W  &= \mathbb{E}_{q(Z_1,Z_2|X_1,X_2,B)}[\log p(B|Z_1,Z_2)]- KL[q_1(Z_1|X_1,B)||p_1(Z_1)]\notag\\
 &-KL[q_2(Z_2|X_2,B)||p_2(Z_2)]+ \delta RFF\; HSIC(\mu_1,S)
\end{align}
where $\delta$ is a hyperparameter, $KL$ is the Kullback-Leibler divergence, and $p_1,p_2$ are Gaussian priors for $Z_1$ and $Z_2$. This method can also be extended to the case where we also seek independence between $Z_2$ and another protected variable.
To have a better understanding of the methodology, we can draw a parallel in \ref{linear embedding} with a linear embedding for data that are elements of a real vector space. We can reformulate the problem of principal component analysis under independence constraint as an optimization
problem with a supplementary term that encourages the independence.

\section{Model adaptation to the Spipoll data set}
\label{Spipoll section}
\subsection{Paradigm shift}

A common practice in ecology to study plant-pollinator interactions is to consider plant and insect species as nodes of a bipartite network, with edges determined by the interactions observed between the two. In the case of Spipoll data set, this implies aggregating all observations of the interactions between a given plant and insect species but also all the covariates describing observation conditions, such as date and climate.

Such aggregation is not straightforward 
so we propose to change paradigm by considering a bipartite network where the first type of nodes is the session of observations, and the second type corresponds to insects observed during the session. Each session has the previously mentioned covariates and a one-hot encoding describing the plant genus. 
This paradigm allows directly using the Spipoll data without doing any aggregation.
Link prediction task in this situation aims to predict which insect will be present during a given observation session. However, we still wish to ultimately obtain a bipartite plant-insect network, as this is the most widely used tool in this field of study. To ensure that the latent space could also be used to create a plant-insect network, we propose two methods to generate the corresponding plant-insect network, using the same decoder.
We propose in our setting to define $S$, the protected variable, as the number of participation from the user. This number of participation would work as a proxy of the user's experience. By employing this measure, we aim to construct a latent space that remains unaffected by variations in observers' experience levels.

\subsection{Model}
In this part, we elaborate on the application of our methodology, taking into account the specific requirements of the Spipoll data set.
We consider $B$ our $n_1\times n_2$ incidence matrix, where the $n_1$ rows correspond to the number of sampling session, and the $n_2$ columns correspond to the number of different observed pollinators in the dataset. For all $i$ and $j$ , $B_{i,j} \in \{0,1\}$ describes the absence or the presence of the pollinator $j$ during the session $i$. 
Let $P = (P_{i,k}) $, $i=1,\dots,n_1,\ k=1,\dots,u$ where $u$ corresponds to the number of observed taxa of plants. $P_{i,k}\in \{0,1\}$ is a binarized categorical variable that describes the plant taxonomy of the $i^{th}$ session. For all $i$, there is only one coordinate $k$ such that $P_{i,k}=1$ while the others are equal to 0. To build the $u \times n_2$ binary adjacency matrix $B'$ of plant-pollinator interactions from the session-pollinator matrix $B$,we check for each plant $k$ and each pollinator $j$ if there are any session $i$ where $P_{i,k}= 1$ and $B_{i,j}=1$. Should it be the case, then $B'_{k,j} =1$, otherwise $B'_{k,j}=0$. This can be computed using $B_{k,j}' =\mathbf{1}\{ (P^{\top}B)_{k,j}>0\}$, where $\mathbf{1}$ is the indicator function.  

We create a second latent space from a realization of the first:
 
 $$\Tilde{q}(Z'_1|Z_1,P) = \prod_{l=1}^{u} \Tilde{q}_1(Z'_{1l}|Z_1,P)$$
 
 where $Z'_1$ is a $u \times D$ matrix, the rows of which $(Z'_{1l}\in\mathbb{R}^D)_{1\le l\le u}$ are the latent variables associated to each plant.
 For any $l$, $\Tilde{q}_1(Z'_{1l}|Z_1,P)$ samples uniformly one $Z_{1i}$ among the ones where $P_{i,l}=1$.
We use the same decoder for both latent space.

It is also possible to estimate $B'$ from the reconstruction $\widehat{B}$ itself, by averaging all the predicted probabilities of interaction by plants. This average is estimated using $\Tilde{P}_{i,k}= \frac{P_{i,k}}{\sum_{l=1}^{n_1}P_{l,k}} $ and $\widehat{B'} = \Tilde{P}^\top \widehat{B}$.

\begin{figure}[H]
   
\adjustbox{scale=1,center}{%
\begin{tikzcd}
B,X_1,X_2 \arrow[r,"q"] \arrow[d, "\mathbf{1}\{ (P^{\top}B)>0\}"] & Z_1, Z_2 \arrow[r,"p"] \arrow[dd,"\Tilde{q}"] & \widehat{B} \arrow[d,"\Tilde{P}^T \widehat{B} ",dashed] & \text{AUC}(\widehat{B},B) \\
B' &  & \widehat{B'} & \text{AUC}(\widehat{B'},B') \\[-25pt]
& Z'_1,Z_2  \arrow[r,"p"]& \Tilde{B'}& \text{AUC}(\Tilde{B'},B')  
\end{tikzcd}
}
\caption{Summary of the model used for the training of the Spipoll data set}
\label{diagram}
\end{figure}
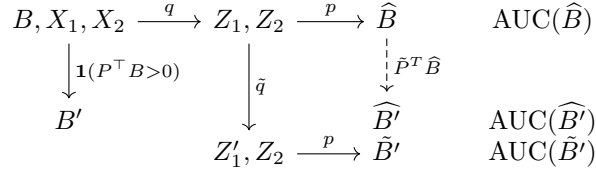

The protected variable $S$ is the log base 10 of the number of observation session the user has already performed. 
 To fit this model, we minimize the same loss as in \cref{loss1} with a supplementary term
 
 $$L'_W = L_W + \mathbb{E}_{\Tilde{q}(Z'_1,Z_2|X_1,X_2,B,P)}[\log p(B'|Z'_1,Z_2)].$$

\section{Application on simulated data}

This  simulation study tries to replicate numerically the sampling process taking place in the Spipoll data set, and to see if our method can yield better result than not taking into account the user's experience. ,We also compare our methodology with an adversarial learning algorithm (ADV) \citep{zhangMitigatingUnwantedBiases2018} where the output $\mu_1$ is then used as an input
to a 4-layer perceptron, which attempts to
predict the protected variable $S$. The loss is then penalized if the predicted output is correlated with the protected variable. 
An application of the fair and bipartite graph variational auto-encoder on a simple simulated case is available in the supplementary materials, \ref{SIMULATION}.
\subsection{Settings}

\begin{figure}[t]
    \centering
    \includegraphics[width=0.95\textwidth]{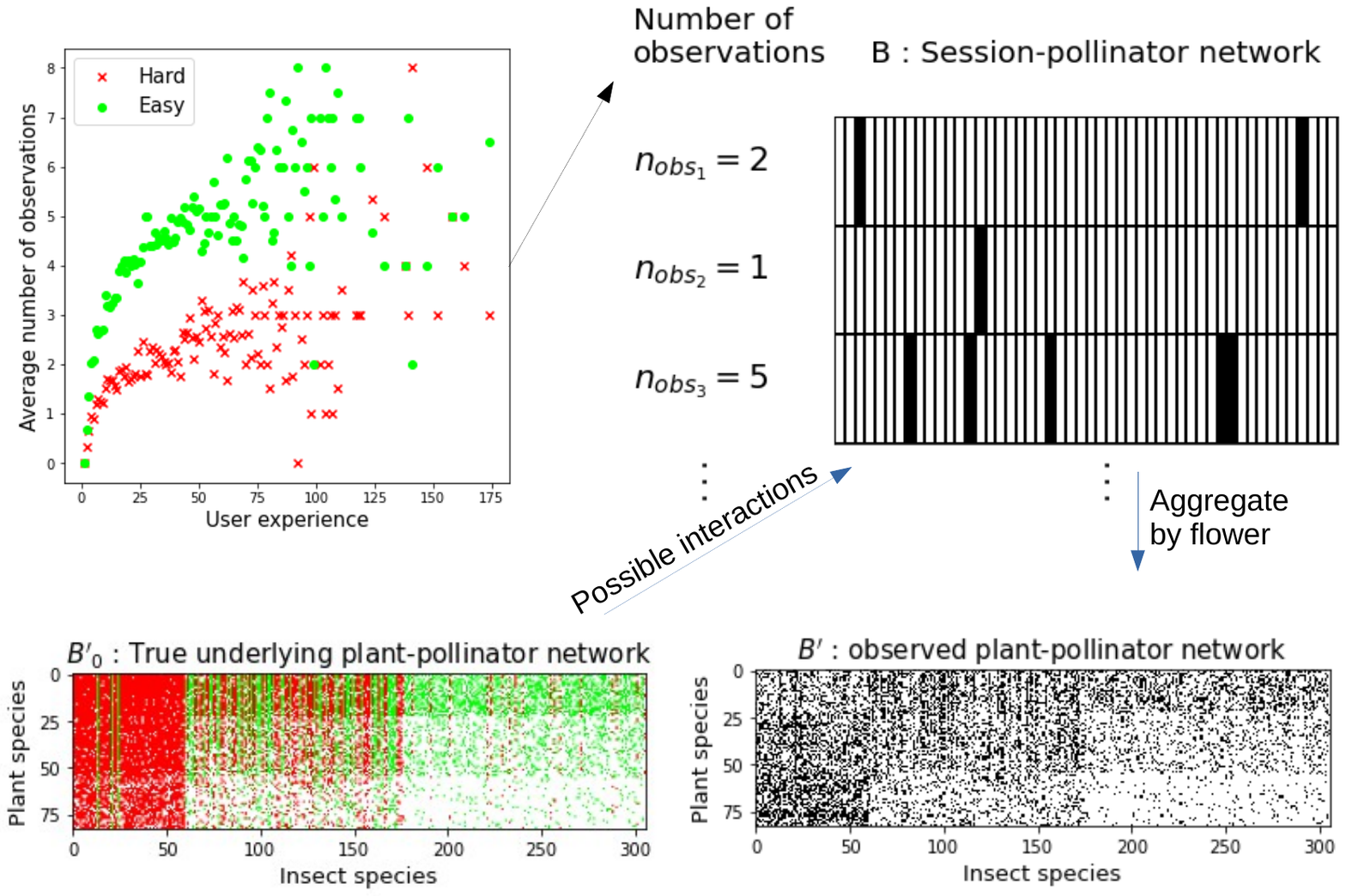}
    \caption{Numerical replication of the sampling process. We simulate various level experience, which describes how many insect the user will observe during a session (top left). During the session, the user can only observe interactions given by one randomly selected row of $B'_0$ (bottom left), with probability depending on the difficulty. The observed insect are reported in the session-pollinator network (top right), which give us $B$. By aggregating this network by plants, we then have the observed network $B'$ (bottom right).}
    \label{sampling_simulation}
\end{figure}

\textbf{Underlying plant-pollinator network : } An underlying plant-insect network $B_0'$ is generated in order to account for possible interactions. It consists of a bipartite SBM made of $u=83$ plants and $n_2= 306$ insects, with parameters $$\alpha =  (0.3,0.4,0.3),\quad \beta= (0.2,0.4,0.4),\quad  \pi = \begin{bmatrix} 0.95 & 0.80 & 0.50\\ 0.90 & 0.55 & 0.20\\  0.70 & 0.25 & 0.06 \end{bmatrix}, $$ 

where $\alpha$ is the row groups proportion, $\beta$ the columns group proportion $\pi$ denotes the connectivity matrix.
This means that for each plant $k$ (resp. insect $j$), there is a latent variable $V^1_k \in \{1,2,3\}$ (resp. $V^2_j \in \{1,2,3\}$) such as  $V^1_k$ follows a multinomial distribution $Mult(1,\alpha)$, $V^2_j \sim Mult(1,\beta)$ and the probability of having an interaction between plant $k$ and insect $j$ is given by $\mathbb{P}(B_{0 k,j}'=1 |V^1_k,V^2_j) = \pi_{V^1_k,V^2_j}$. One simulation of this network can be seen in the bottom left part of \cref{sampling_simulation} . The given parameters  correspond to a nested network, a model often encountered in ecological study.
We also suppose that the insects are separated in two groups :  ``easily observable" ($h_j=easy$) and ``hardly observable" ($h_j=hard$). For the simulation, we assume that the difficulty to be observed depends on the latent variable $V^2_j$, such as $\mathbb{P}(h_j = hard)=1 - \mathbb{P}(h_j = easy) = H_{V^2_j}$ with $H = (0.9,0.6,0.1)$.
Easily observable insects are given a weight $w_{easy} = 0.8 $ and  hardly observable insects are given a weight $w_{hard}=0.2$. 

\noindent\textbf{Session-pollinator network : }
Let $n_1 =3000$ be the number of observers, we suppose for the simulation that their user's experience is given for $1\leq i \leq n_1$ by $S_i = round(S'_i) +1,  $ where $S'_i \sim \varepsilon(21)$, with $\varepsilon(\lambda)$ an exponential random variable of density $f_{\lambda}(x) = \frac{1}{\lambda}e ^{-\frac{x}{\lambda}} \mathbf{1}_{\{x\geq0\}}$. Each user will select uniformly at random one flower $k$, and will observe $obs_i = round(2\log(S_i))$ possible interactions from the $k$-th row of the matrix $B_0'$ at random, with a probability proportional to the difficulty, given by the weight $w_{easy}$ and $w_{hard}$. Once the observations-insects network is constructed, we also have access to the observed plant-pollinator network. The covariates related to the observation sessions are defined by $X_1 = P$ where $P$ is the binarized categorical variable (83 columns) describing the observed plant genus. No feature is considered for the pollinators, therefore $X_2$ is set as the identity matrix of $n_2$ rows, as done in \cite{Kipf}.
A summary of the procedure is available in \cref{sampling_simulation}.

\subsection{Results}

\begin{table}
\caption{Comparison between the Bipartite variational graph auto-encoder and its fair counterparts with simulated data. In the table are reported the mean and standard deviation for 30 trials, except for $\#p_{0.05}$ which is a count.
\label{table_simulation}}
\resizebox{\textwidth}{!}{%
\begin{tabular}{ |c|c|c|c| } 
 \hline
   & BVGAE & fair-BVGAE & ADV \\ 
 \hline
AUC($\widehat{B},B$) & $0.641 \pm 0.013$   & $0.595 \pm 0.021$ & $0.642 \pm 0.014$ \\ 
AUC($\Tilde{B}',B'_0$) & $0.582 \pm 0.017$   & $0.592 \pm 0.024$ & $0.580 \pm 0.021$ \\ 
AUC($\widehat{B'},B'_0$) & $0.613 \pm 0.020$   & $0.655 \pm 0.040$ & $0.608 \pm 0.026$ \\ 
HSIC & $2.86\times 10^{-3} \pm 0.27\times 10^{-3}  $  & $1.32\times 10^{-5}\pm 0.31\times 10^{-5} $ & $3.65\times 10^{-3} \pm 0.99\times 10^{-3} $ \\ 
$\#p_{0.05}$ & 30/30     & 0/30 & 30/30 \\ 
$cor$ & $0.380 \pm 0.040 $  & $0.036\pm 0.012$& $0.379 \pm 0.090$  \\ 
$cor_{sp}$($\widehat{B'}$) & $0.467 \pm 0.040 $  & $0.533\pm 0.085$& $0.472 \pm 0.058$  \\ 
 \hline
\end{tabular}
}
\end{table}

The results for the link prediction task for the simulated data set are summarized in \cref{table_simulation}. The predictions were made with the complete dataset $B$, that has been split with 30\% of the edges hidden. 20\% of these hidden edges are used as validation data set, and the remaining 10\% for the test set. Both sets also contain an equivalent amount of non-edges that are not in the train set. Moreover, as we also try to have a latent representation for $B'$, we carefully construct another training set with the edges and non-edges that have already been used for the learning of $B$. The remaining edges and non-edges are used for the test, from which we know the ground truth thanks to the true underlying matrix $B'_0$. We compare the link prediction $\widehat{B}$ at the observation session level (AUC($\widehat{B},B$)), and the link prediction $\widehat{B'}$ and $\Tilde{B'}$ at the plant-insect level (AUC($\widehat{B'},B'_0$), AUC($\Tilde{B'},B'_0$)). \text{AUC}($\Tilde{B'},B'_0$) is calculated with the prediction obtained by the second latent space $Z'_1,Z_2$, whereas AUC($\widehat{B'},B'_0$) is calculated from the prediction obtained by the reconstruction $\widehat{B}$ with $\widehat{B'} = \Tilde{P}^\top \widehat{B}$ (see \cref{diagram}). We also calculate the Spearman correlation $cor_{sp}(\widehat{B'})$ between the predicted matrix $\widehat{B'}$ and the true underlying matrix of probabilities given by the SBM.
Both methods are fit with 1 000 iterations of Adam algorithm with learning rate 0.01, and with a latent space of dimension 4 with $D_+ = D_- = 2$, using a computer equipped with an Intel Xeon(R) CPU E5-1650 v4 and 32GB of RAM. 

From \cref{table_simulation}, we notice a reduction of prediction accuracy for $B$ when enforcing a fair setting in the BVGAE. This is a common consequence of most fairness settings. However,  using fairness in this case has increased the quality of prediction at the plant-pollinator level ($B'$) as we can see with the AUC($\widehat{B'},B'_0$) and the Spearman correlation obtained with the fair-BVGAE. The reconstruction of the plant-pollinator level is clearly better via $\widehat{B'}$
than via $\Tilde{B'}$.

Moreover, the latent space given by the \mbox{BVGAE} is not independent of the protected variable $S$. This can be seen by looking at the p-value of the HSIC independence test and the correlation between $\Tilde{Z_1}$ and $S$. Even if it is not enough to guarantee independence, we can see that the correlation between the latent space and the protected variable is much higher in the BVGAE than in the fair-BVGAE and ADV model. In all the simulations, the independence hypothesis has been rejected for the BVGAE and kept for the fair-BVGAE. The ADV did not manage to have a smaller HSIC than the BVGAE, and the independence hypothesis was rejected most of the time. The ADV model is much harder to calibrate because it requires a second neural network to optimize.

\section{Results on the Spipoll data set}
\label{spipoll_result_section}

\begin{table}[H]
\caption{Comparison between the Bipartite variational graph auto-encoder and its fair counterpart on 10 trials on the Spipoll data set.\label{table2}}
\begin{tabular}{ |c|c|c| } 
 \hline
   & BVGAE & fair-BVGAE \\ 
 \hline
AUC($\widehat{B},B$) & $0.869 \pm 0.003$ & $0.834 \pm 0.007$ \\ 
\rule{0pt}{3ex}
AUC($\Tilde{B'},B'$) & $0.731 \pm 0.004$ & $0.710 \pm 0.016$ \\ 
\rule{0pt}{3ex}
AUC($\widehat{B'},B'$) & $0.777 \pm 0.018$ & $0.758 \pm 0.022$ \\ 
\rule{0pt}{3ex}
 HSIC & $5.33\times 10^{-4} \pm 0.29\times 10^{-4}  $  & $1.48\times 10^{-6}\pm 0.22\times 10^{-6} $ \\ 
\rule{0pt}{3ex}
$\#p_{0.05}$ & 10/10     & 0/10 \\ 
$cor$ & $0.1 \pm 0.006 $  & $0.003\pm 0.001$ \\ 
 \hline
\end{tabular}
\end{table}


\begin{figure}[h!]
\centering
\includegraphics[width=0.74\textwidth]{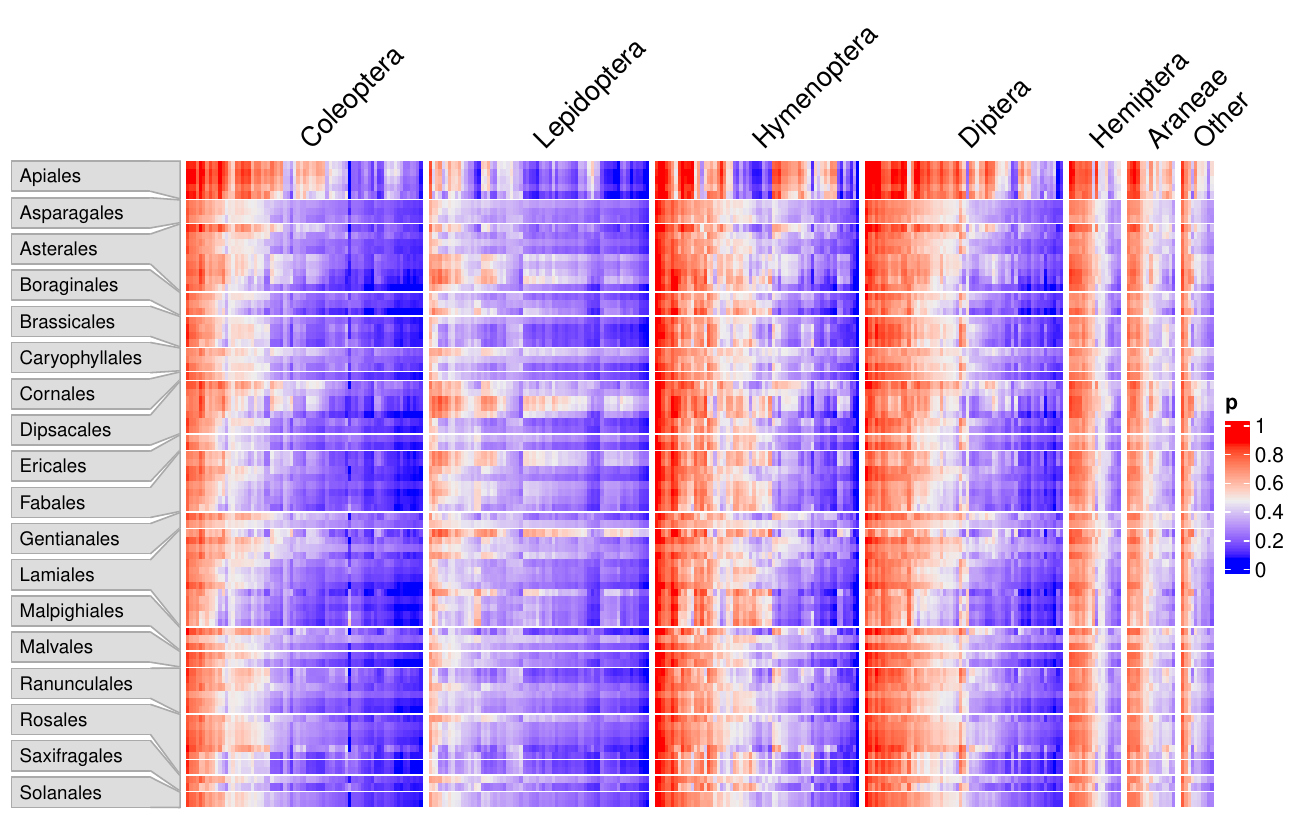}
\includegraphics[width=0.74\textwidth]{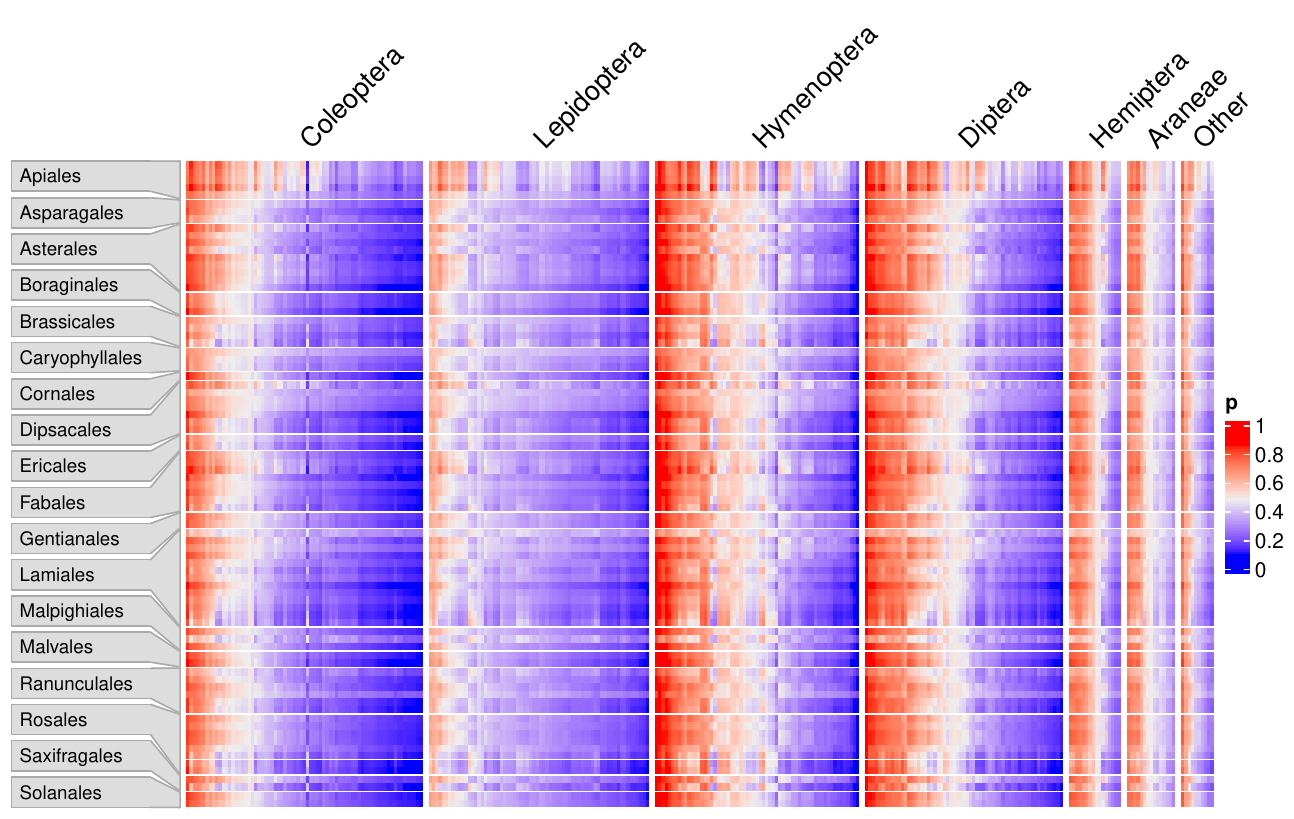}
\caption{Estimated probabilities $\widehat{B'}$ of connection between plants and insects on the Spipoll data set obtained with BVGAE (top) and the fair-BVGAE  (bottom). Each row and column represent respectively a genus of plant and insect, which have been grouped by taxonomic orders. 
}
\label{spipoll_result_connect}
\end{figure}
We consider the observation period of the Spipoll data set from 2017 to 2020 included, in metropolitan France and Belgium. We consider a total of $n_1=12754$ observation sessions, where $n_2=306$ taxa of insects and $u=83$ genera of plants have been observed. The covariates related to the observation sessions are $X_1 = [P, T]$ where $P$ is the
binarized categorical variable (83 columns) describing the observed plant genus, concatenated with $T$, the average temperature on the day at the observation location, provided by the European Copernicus Climate data \citep{cornesEnsembleVersionEOBS2018}. No feature is considered for the pollinators, therefore $X_2$ is set as the identity matrix with $n_2$ rows. 
We use the same training setting as previously, with $D_+ = D_- = 2$. We justify this choice by looking at the estimated mean of $AUC(\widehat{B})$ for different numbers of dimensions $D_+$ and $D_-$ in \ref{spipoll_appendix}. Contrary to the simulation study, the underlying plant-pollinator interaction matrix $B'_0$ is not available. Therefore, the AUC for the prediction of $B'$ are performed on the observed plant-pollinator network which is incomplete.

Looking at the results in \cref{table2}, both methods have better prediction on $B$ than on $B'$, and calculating $\widehat{B'} = \Tilde{P}^\top \widehat{B}$ yields better results than calculating $\Tilde{B'}$. The BVGAE has better AUC than its fair-counterpart, however, even if the linear correlation between the latent space and the protected variable $S$ seems low (0.051), the latent space of the BVGAE is not independent of $S$ according to the HSIC test. Although the AUC($\widehat{B}$) decreased in average from 0.869 to 0.834, the fair-BVGAE has a latent space independent of $S$, which is the target result: the latent representation $Z_1$ is independent of the users' experience levels. This can yield better prediction as seen in our simulation study, but it is difficult to assert because the true underlying network is inaccessible.  Looking at \cref{spipoll_result_connect} we can see a change of structure when accounting for sampling bias leading to a change of our ecological understanding of this plant-pollinator network. For example,
the fair adjustment overall increases the probability of connection, notably of Lepidoptera revealing a higher contribution of butterflies to pollination.  
A checkerboard structure is observed for the interactions of Hymenoptera in the BVGAE 
suggesting contrasted preferences among plants and pollinators. This structure is smoothened in the fair-BVGAE
indicating that such differences in preferences have no ecological ground but are related to the sampling process.

\begin{figure}[t]
\centering
\includegraphics[width = 0.8\textwidth]{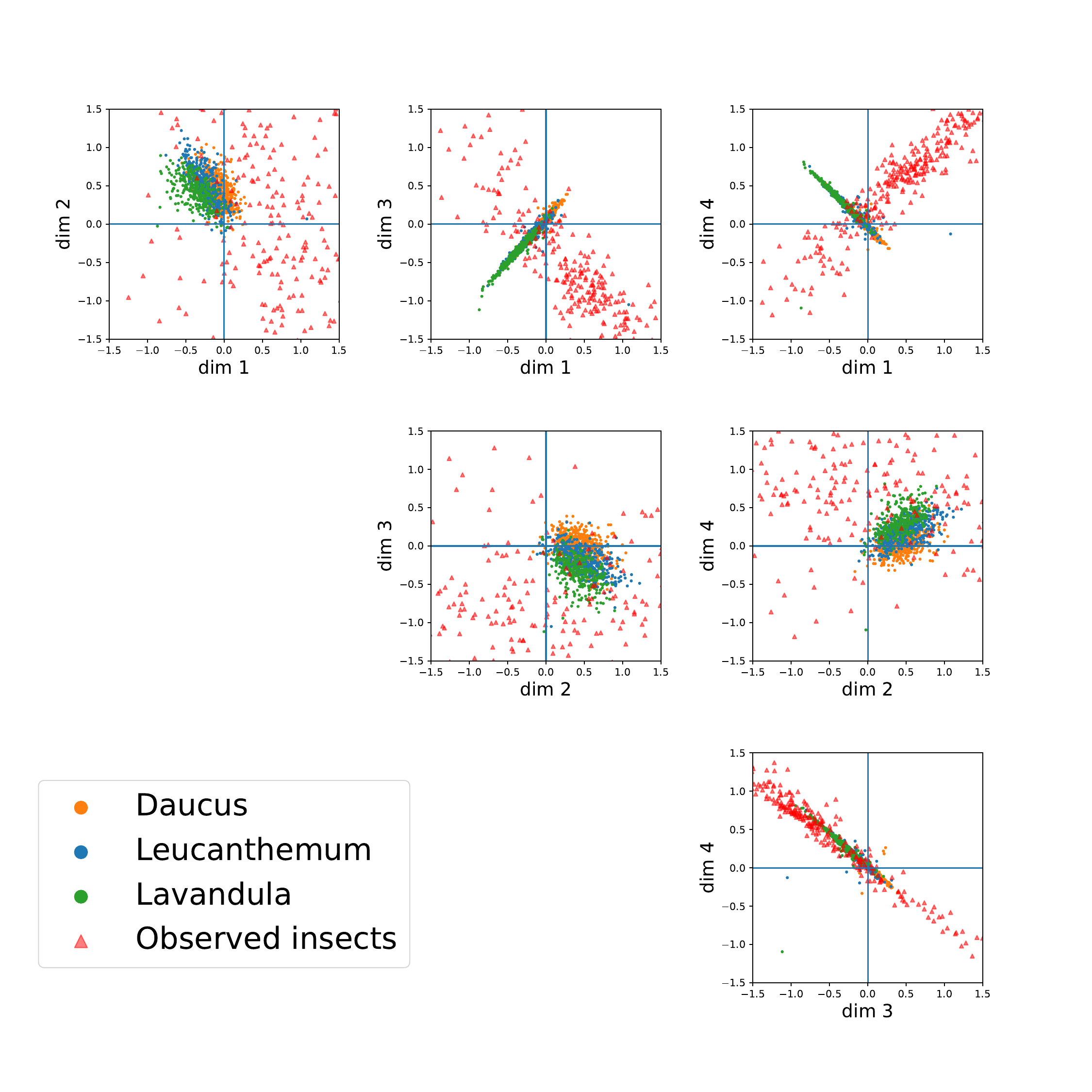}
\caption{Focus on the embeddings provided 
by the fair-BVGAE for the
 observation sessions performed
on the genera \textit{Daucus}, \textit{Leucanthenmum} and \textit{Lavandula}.
}
\label{Daucus}
\end{figure}
In addition to the plant-pollinator network, the method provides a session-pollinator embedding. In order to analyze the pattern of plant pollinator interaction, we can focus on the sessions where a particular genera has been observed, in order to have a closer look on specific interactions or to compare interaction profiles. As an example
in \cref{Daucus}, we focus on three plant genera, Daucus, Leucanthemum and Lavandula. We remind that 
the
method provides an embedding of dimension $D = D_+ + D_- = 4$ with $D_+ = D_- = 2$, which means that for the first two dimensions, insects and sessions that are embedded in the same direction are more likely to be connected, and the ones in the opposite direction are less likely to be connected. On the contrary, insects and sessions that are embedded in the same direction for the third and fourth dimensions are less likely to be connected, while the ones in the opposite direction are more likely to be connected. In the top-left panel of \cref{Daucus}, sessions conducted on Daucus are positioned further to the right along the first axis compared to those conducted on Leucanthemum or Lavandula. Since most observed insects are also located on the right side of the plot, this suggests a higher probability of insect observations during sessions on Daucus than on the other genera and thereby that it is a more generalist plant than the two others. This can also be seen in the bottom-right panel, where sessions conducted on Lavandula are positioned in the same direction as most insects along the third and fourth axis, which suggests that Lavandula is more specialized than the other two genera in terms of plant–pollinator interactions. Representations of latent spaces with all plant genera are available in \ref{original}.

\section{Conclusion}
In this paper, we proposed not only a bipartite extension of the graph variational auto-encoder, but also a new method to have a fair latent representation with respect to a continuous variable. 
We handled a particular structure of data: observed 
collections of pollination events with numerous covariates from
which are derived a plant-pollinator interaction network. We then used our model to tackle the sampling effects affecting this large ecological network, an issue of particular importance when the sampling involves citizen science.
Even if our proposed model is specific to the study of Spipoll, our contribution using GNN and the HSIC as a bias mitigation strategy can be adapted to provide interesting results in other fields of study where both networks and sampling bias are at stake. 
A question of interest
that we left for future research is to 
assess the effects of the covariates on the pollination interactions.

\paragraph{Acknowledgments}
This work was partially supported
by the grant ANR-18-CE02-0010-01 of the French National Research Agency ANR (project EcoNet).
Emre Anakok was funded by the MathNum department of INRAE.

\bibliography{biblio.bib}

\clearpage
\appendix

\section{The linear embedding case}
\label{linear embedding}
\subsection{}
Let $X$ a $n\times d$ matrix and let $S$ be $n \times d_s$ matrix. Without loss of generality, we assume that each column of $X$ has been centered. We wish to perform a one dimensional principal component analysis on $X$ that would yield us a vector $v$ and a lower dimensional embedding of $X$ given by $Xv$ that maximizes the variance. However, we wish to have a latent representation $Xv$ independent of the protected variable $S$.  
If we were in the context of probabilistic PCA \citep{tippingProbabilisticPrincipalComponent1999} where $X$ and $S$ would have been multivariate Gaussian, 
projecting $X$ onto the space orthogonal to $S$: $S^\perp$ beforehand would have been enough to guarantee the independence between $S$ and the latent representation $Xv$, this can be solved using PPCA with covariates \citep{kalaitzisResidualComponentAnalysis2012}. 
We show that this approach is equivalent to find the optimal projection with respect to an independence constraint.

We note $P_S = S(S^\top S)^{-1}S^\top$ the orthogonal projection on the span of $S$ and  $P_{S^\bot}=I_d - P_SX$ the orthogonal projection on the space orthogonal to the span of $S$. 

\paragraph{Proposition 1}
Assume that X is centered. Let $\Lambda = \frac{1}{n} X^\top X$. Assume that $X$ and $S$ are jointly Gaussian. The solution of the maximization problem given by
the following Lagrangian:
\begin{equation}\nonumber
    L(v) = v ^\top \Lambda v + \lambda_1 (1-v^\top v) + \lambda_2 ||S^\top Xv||^2
\end{equation}

can also be obtained by computing the first component of the PCA of $P_{S^\bot}X$. 
\paragraph{Proof}

We can see that
$$ X^\top X = (P_SX + P_{S^\bot}X)^\top (P_SX + P_{S^\bot}X) =  (P_SX)^\top P_SX +  (P_{S^\bot}X)^\top P_{S^\bot}X$$

and $$ ||S^\top Xv||^2 =v^\top X^\top S  S^\top Xv,$$

thus, we have

$$L = \frac{1}{n}v ^\top(P_SX)^\top P_SXv + \frac{1}{n} v ^\top(P_{S^\bot}X)^\top P_{S^\bot}Xv + \lambda_1 (1-v^\top v) + \lambda_2 ||S^\top Xv||^2 .$$

Derivating the Lagrangian yields 

\begin{equation}
   \frac{\partial L}{\partial v} =  \frac{2}{n}(P_SX)^\top P_SXv + \frac{2}{n} (P_{S^\bot}X)^\top P_{S^\bot}Xv - 2\lambda_1 v + \lambda_2 2 X^\top S  S^\top Xv  = 0 ,
   \label{eq1}
\end{equation}

\begin{equation}\nonumber
    \frac{\partial L}{\partial \lambda_1} = 1-||v||^2  =0,
\quad
    \frac{\partial L}{\partial \lambda_2} = ||S^\top Xv||^2 =0 .
\end{equation}

First, we can see that 
$$||S^\top Xv||^2 =0  \implies S^\top Xv = 0 $$ which allows us to plug in Equation \eqref{eq1}:  

\begin{equation}
     \frac{\partial L}{\partial v} =  \frac{2}{n}(P_SX)^\top P_SXv + \frac{2}{n} (P_{S^\bot}X)^\top P_{S^\bot}Xv + 2\lambda_1 v   = 0.
\end{equation}

Moreover,
$$P_S = S(S^\top S)^{-1}S^\top  $$ 
thus
$$P_SXv  = S(S^\top S)^{-1}(S^\top Xv) = 0 .$$

Finally,  \cref{eq1} becomes

$$ \frac{2}{n} (P_{S^\bot}X)^\top P_{S^\bot}Xv - 2\lambda_1 v =0 $$

which is equivalent to search for $\lambda_1$ and $v$ such as 

$$ \frac{1}{n} (P_{S^\bot}X)^\top P_{S^\bot}Xv = \lambda_1 v , $$ in other words, we are looking for the eigenvalues of the covariance matrix of $X$ projected on $S^\top$, which is the same as performing a PCA on  $P_{S^\bot}X$.

\subsection{Comparison of fair linear embedding}
\label{PCA with indep}
Let  $S_i\overset{i.i.d.}{\sim} \mathcal{N}(0,1)$ and $T_i\overset{i.i.d.}{\sim} \mathcal{N}(0,1)$ for $ i \in 1,\dots,n=1000$. Assume that $S \bot T$ and let $Z = (S,T)$. Let $K_{i,j} \overset{i.i.d.}{\sim} \mathcal{N}(0,9)$ be a $2 \times 5$ matrix. Suppose that we observe the $n \times 5$ matrix $X=ZK$  and the protected variable $S$.

We wish to perform a linear embedding 
$$X\xrightarrow[Linear]{} Z\xrightarrow[Linear]{}\widehat{X}$$ 

with three different methods, principal component analysis on $X$, principal component analysis on  $P_{S^\bot}X$, and  principal component analysis on $X$ using the HSIC loss between the latent space and $S$ as an additional loss term. This is the introductory case, where the optimal solution can be obtained with a projection. We aim to investigate if using the HSIC as a loss in this setting would yield a result similar to the optimal one.

\subsubsection{Principal component analysis using $X$}  We consider an encoder $f_{W_0}$ with a one layer neural network of 5 input nodes and 2 output nodes, and a decoder $g_{W_1}$ with 2 input nodes and 5 output nodes. We optimize the weights of the auto-encoder with respect to the mean squared error loss: 
$$\mathcal{L} (W_0,W_1) = \frac{1}{n} ||g_{W_1}(f_{W_0}(X))-X||. $$

\subsubsection{Principal component analysis using $P_{S^\bot}X$}  We consider an encoder $f_{W_0}$ with a one layer neural network of 5 input nodes and 2 output nodes, and a decoder $g_{W_1}$ with 2 input nodes and 5 output nodes. We optimize the weights of the auto-encoder with respect to the mean squared error loss: 
$$\mathcal{L} (W_0,W_1) = \frac{1}{n} ||g_{W_1}(f_{W_0}(P_{S^\bot}X))-X|| .$$

The difference between the precedent model is that the encoder takes as input $P_{S^\bot}X$. In our case, this would erase all effect from the protected variable in the latent space.

\subsubsection{Principal component analysis using $X$ and the HSIC loss}  We consider an encoder $f_{W_0}$ with a one layer neural network of 5 input nodes and 2 output nodes, and a decoder $g_{W_1}$ with 2 input nodes and 5 output nodes. We optimize the weights of the auto-encoder with respect to the MSE and HSIC loss: 
$$\mathcal{L} (W_0,W_1) = \frac{1}{n} ||g_{W_1}(f_{W_0}(X))-X|| + \delta RFF\: HSIC(f_{W_0}(X),S).$$ Here we have chosen $\delta = 10^5$. 

For all the presented method, we fit the weights using 200 steps of the Adam algorithm with learning rate 0.01. For the HSIC loss, we fit the  algorithm 10 times with different initialization before selecting the one with the lowest HSIC value.
We then simulate a test set of 200 observations following the same probability law than the training test. Mean squared error, HSIC, number of time the independence hypothesis is rejected, and the Euclidean norm of the covariance between the latent space and the protected variable $S$ are reported in Table \ref{table_acp}

\begin{figure}[h]
  \includegraphics[width=0.30\textwidth]{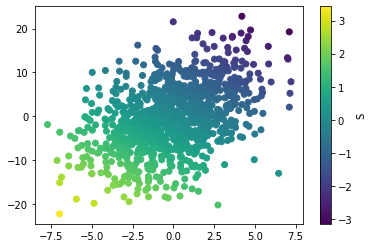}
  \includegraphics[width=0.30\textwidth]{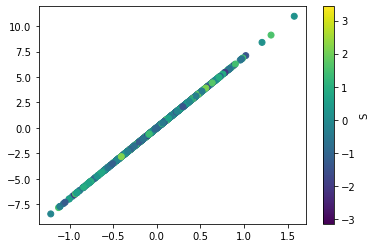}
  \includegraphics[width=0.30\textwidth]{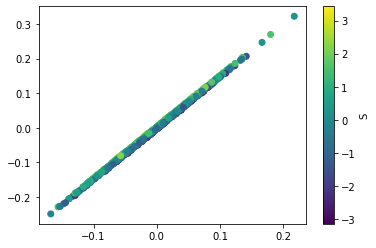}
  
  \includegraphics[width=0.30\textwidth]{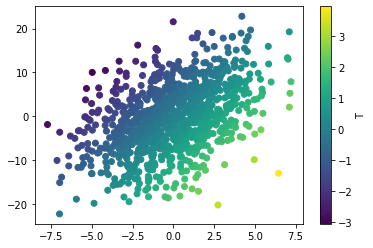}
  \includegraphics[width=0.30\textwidth]{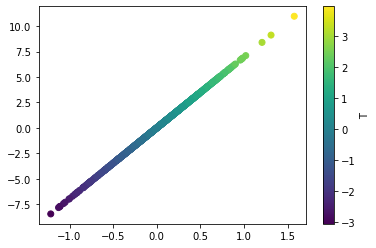}   
  \includegraphics[width=0.30\textwidth]{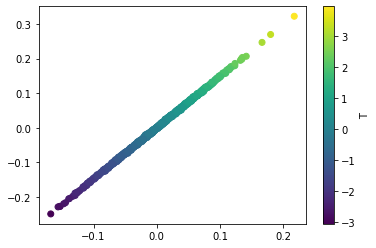}
    \caption{Latent representation of the linear embedding performed on $X$ (left), performed on $P_{S^\bot}X$ (middle), and performed on $X$ with the HSIC loss (right). The latent space is colored according to the protected variable $S$ (top) and the variable $T$ (bottom). Adding the HSIC as a loss term yielded similar latent space as making the embedding on $P_{S^\bot}X$, except that the points are not perfectly aligned. }

\label{latent_PCA}
\end{figure}

\begin{table}
\small
\caption{Comparison between the PCA, the projected PCA and the HSIC PCA}
\label{table_acp}
\resizebox{\textwidth}{!}{%
\begin{tabular}{ |c|c|c|c| } 

 \hline
   & PCA & PCA with projection & PCA with HSIC\\ 
 \hline
\rule{0pt}{2.5ex}MSE & $7.16\times 10^{-2} \pm 21.4\times 10^{-2}  $ & $9.88 \pm 6.34$  & $8.80 \pm  5.54$\\ 

\rule{0pt}{3ex} HSIC & $3.15\times 10^{-2} $  & $1.15\times 10^{-3}  $ & $2.21 \times  10^{-3} $\\ 
&$\pm 1.02\times 10^{-2} $ &$\pm 0.60\times 10^{-3} $ &$\pm 4.81\times 10^{-3} $ \\

$\rule{0pt}{3ex}\#p_{0.05}$ & 100/100     & 3/100 & 10/100\\ 
$cor$ & $0.991 \pm 0.229 $  & $6.83\times 10^{-2} \pm 4.92\times 10^{-2} $ & $0.142 \pm 0.241$\\ 
 \hline
\end{tabular}
\label{table_PCA}
}
\end{table}

Removing the protected variable $S$ from the original data $X$ has increased the MSE, which was the expected behavior because $X$ depends on $S$. However, doing the projection or adding the HSIC as a penalty term in the loss have yielded results where the latent space became independent of the protected variable. Looking at \cref{latent_PCA} we can see an example where the latent space of the PCA with projection and the PCA with HSIC are similar.

\section{Extensive simulation study}
All the simulations are reproducible and available on \url{https://github.com/AnakokEmre/Bipartite-and-fair-VGAE}.
\subsection{Simulation in the simple case}
\label{SIMULATION}
\subsubsection{Setting}

In this simulation, we are going to generate bipartite networks made of $n_1=1000$ rows and $n_2 =100$ columns. 
Let $S_i\overset{i.i.d.}{\sim} \mathcal{N}(0,1)$ for $i = 1,\dots,n_1$ and $T_i \overset{i.i.d.}{\sim} \mathcal{N} (0,1)$ for $i = 1,\dots,n_1$ and independent of $S$. We suppose that $S$ is the protected variable. Let $Z_1 = (S,T) \in \mathbb{R}^{n_1 \times 2}$ be the 2-column matrix made with both $S$ and $T$. Let $Z_2\overset{i.i.d.}{\sim} \mathcal{N}\left(\begin{bmatrix} 0\\0 \end{bmatrix}, \begin{bmatrix} 1 & 0 \\ 0 & 1 \end{bmatrix}\right)\in \mathbb{R}^{n_2 \times2}$.
We simulate our bipartite adjacency matrix with a Bernoulli distribution $ B_{i,j} \overset{i.i.d.}{\sim} \mathcal{B}(sigmoid(z_{1i}^\top\mathbf{I}_{D_+,D_-}z_{2j}))$.

First, we fit a classical bipartite and variational graph auto-encoder on $B_{i,j}$. We expect that this auto-encoder would yield a latent representation $\Tilde{Z_1}$ correlated with $S$ and $T$. 
We then fit our bipartite and fair auto-encoder to compare the results and see if the yielded latent space is independent of $S$. We also compare our methodology with an adversarial learning algorithm (ADV) \citep{zhangMitigatingUnwantedBiases2018} where the output $\mu_1$ is then used as an input
to a 4-layer perceptron, which attempts to
predict the protected variable $S$. The loss is then penalized if the predicted output is correlated with the protected variable.

\subsubsection{Results}
\label{simulation_result}
\begin{figure}
\centering
\includegraphics[width=0.6\textwidth]{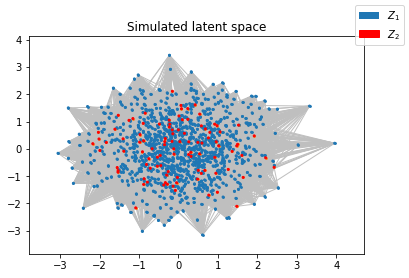}
\caption{Simulated latent space for generating bipartite network $B_{i,j}$. $Z_1 = (S,T)$ is represented in blue. $Z_2$ is represented in red and is independent of $Z_1$. The probability of connection between the node $i$ and $j$ will increase as the distance between their latent representation decreases. }
\label{fig1}
\end{figure}

The results for the link prediction task in the simulated network are summarized in \cref{table1}.
The simulations were done with dataset splits, with 30\% of the edges hidden. 20\% of these hidden edges are used as validation data set, and the remaining 10\% for the test set. Both sets also contain an equivalent amount of non-edges that are not in the train set. We compare the methods with the area under the ROC curve (AUC) score, the $HSIC$ between the latent space $\Tilde{Z_1}$ and $S$, 
the number of times the p-value associated with the HSIC independence test is lower than $0.05\%$ ($\#p_{0.05}$) and the Euclidean norm of the correlation matrix between $\Tilde{Z_1}$ and $S$ ($cor$).
In the table, are reported the mean and standard deviation for 100 trials, except for $\#p_{0.05}$ which is only a count.
We set the hyperparameter $\delta = n_1$.
For each trial, the simulations begin with 10 random initializations, and were fit using 1000 iterations of the Adam algorithm with learning rate $ 0.01$, using a computer equipped with an Intel Xeon(R) CPU E5-1650 v4 and 32GB of RAM. The model that achieved the most favorable performance on the validation test set is then selected to evaluate the performance on the test dataset.

\begin{table}
\caption{\label{table1}
Comparison between the Bipartite variational graph auto-encoder and its fair counterparts on 100 trials with simulated data.}
\centering
\begin{tabular}{ |c|c|c|c| } 
\hline
   & BVGAE & fair-BVGAE & ADV\\ 
 \hline
\rule{0pt}{2.5ex}AUC & $0.753 \pm 0.013$ & $0.664 \pm 0.014$  & $0.668 \pm  0.036$\\ 

\rule{0pt}{3ex} HSIC & $0.041 \pm 0.002$   & $2.36\times 10^{-6} \pm 1.18\times 10^{-6} $ & $1.57 \times  10^{-3} \pm 3.21 \times  10^{-3} $ \\ 
$\rule{0pt}{3ex}\#p_{0.05}$ & 100/100     & 0/100 & 81/100\\ 
$cor$ & $0.940 \pm 0.022 $  & $0.009\pm 0.006$ & $0.12 \pm 0.195$\\ 
 \hline
\end{tabular}
\end{table}

\begin{figure}[h!]
\centering
\includegraphics[width=.45\textwidth]{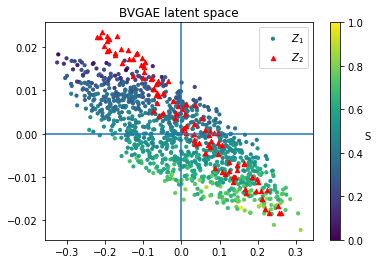}
\includegraphics[width=.45\textwidth]{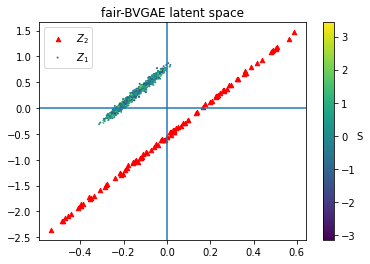}

\caption{Estimated latent space for the bipartite variational graph auto-encoder (left) and the fair bipartite variational graph auto-encoder (right). }

\label{fig2}
\end{figure}

As expected in a fairness setting, the AUC for link prediction decreases when we penalized the reconstruction with the HSIC, because in our case, $S$ is directly related to the probability of connection between the nodes. However, the latent space given by the \mbox{BVGAE} is not independent of the protected variable $S$. This can be seen by looking at the p-value of the HSIC independence test and the correlation between $\Tilde{Z_1}$ and $S$. Even if it is not enough to guarantee independence, we can see that the correlation between the latent space and the protected variable is much higher in the BVGAE than in the fair-BVGAE and ADV model. However, in all the simulations, the independence hypothesis has been rejected for the BVGAE and kept for the fair-BVGAE. The ADV model managed to have a smaller HSIC than the BVGAE, however the independence hypothesis was rejected most of the time. The ADV model is much harder to calibrate because it requires a second neural network to optimize.

An example of the latent space of BVGAE and \mbox{fair-BVGAE} can be seen in figure \ref{fig2}. Looking at the coloring, we can see for the BVGAE that the latent space is clearly correlated with $S$ , while the latent space of the fair-BVGAE does not share structure with the protected variable $S$. The HSIC test between the fair latent space and $S$ yields us a p-value equals to $0.139$, we do not reject the hypothesis that the latent space $Z_1$ is independent of $S$. {Simulation with binary protected variable is available in \ref{binary}.}

\subsection{Impact of hyperparameter $\delta$} We remind the expression of the variational loss given in  \cref{loss1}: 
\begin{align}
 L  &= \mathbb{E}_{q(Z_1,Z_2|X_1,X_2,B)}[\log p(B|Z_1,Z_2)]- KL[q_1(Z_1|X_1,B)||p_1(Z_1)]\notag\\
 &-KL[q_2(Z_2|X_2,B)||p_2(Z_2)]+ \delta RFF\; HSIC(\mu_1,S)\notag.
\end{align}
In this expression, $\delta$ is the hyperparameter associated with the $RFF\: HSIC$. 
Setting $\delta = 0$ yields the same result as fitting the classical BVGAE. The following simulation study is performed to study the impact of this hyperparameter on the different scores.

\subsubsection{Setting}
The settings are nearly identical as in \ref{SIMULATION}.
In this simulation, we are going to generate bipartite networks made of $n_1=1000$ rows and $n_2 =100$ columns. 
Let $S_i\overset{i.i.d.}{\sim} \mathcal{N}(0,1)$ for $i = 1,\dots,n_1$ and $T_i \overset{i.i.d.}{\sim} \mathcal{N} (0,1)$ for $i = 1,\dots,n_1$ and independent of $S$. We suppose that $S$ is the protected variable. Let $Z_1 = (S,T) \in \mathbb{R}^{n_1 \times 2}$ be the 2-column matrix made with both $S$ and $T$. Let $Z_2\overset{i.i.d.}{\sim} \mathcal{N}\left(\begin{bmatrix} 0\\0 \end{bmatrix}, \begin{bmatrix} 1 & 0 \\ 0 & 1 \end{bmatrix}\right)\in \mathbb{R}^{n_2 \times2}$. We simulate our bipartite adjacency matrix with Bernoulli  $ B_{i,j} \overset{i.i.d.}{\sim}  \mathcal{B}(sigmoid(z_{1i}^\top\mathbf{I}_{D_+,D_-}z_{2j}))$.

We fit the fair-BVGAE with the variational loss $\mathcal{L}$ with hyperparameter $\delta \in$ \{0, 10, 100, 200, 500, 1000, 2000\}.  
\subsubsection{Results}

The results for the link prediction task in the simulated network are summarized in \cref{table_delta}.
The simulations were done with  dataset splits, with 30\% of the edges hidden. 20\% of these hidden edges are used as validation data set, and the remaining 10\% for the test set. Both sets also contain an equivalent amount of non-edges that are not in the train set.
In the table are reported the mean and standard deviation for 100 trials, except for $\#p_{0.05}$ which is only a count.

For each trial, the simulations begin with 10 random initialization, and were fit using 1000 iterations of the Adam algorithm with learning rate $ 0.01$.  The model that achieved the most favorable performance on the validation test set is then selected to evaluate the performance on the test dataset.

This procedure is repeated on the same network for each value of $\delta$.

\begin{figure}[h]
  \includegraphics[width=0.5\textwidth]{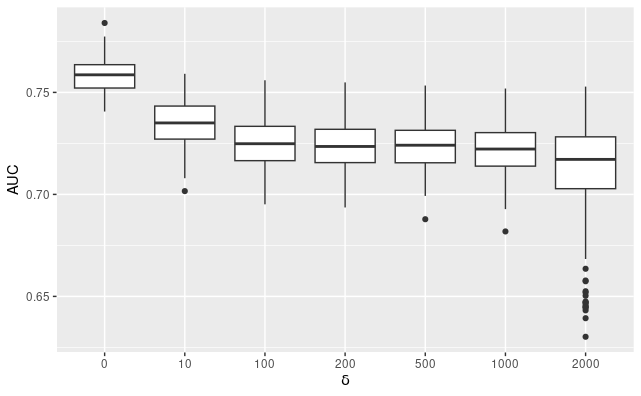}
  \includegraphics[width=0.5\textwidth]{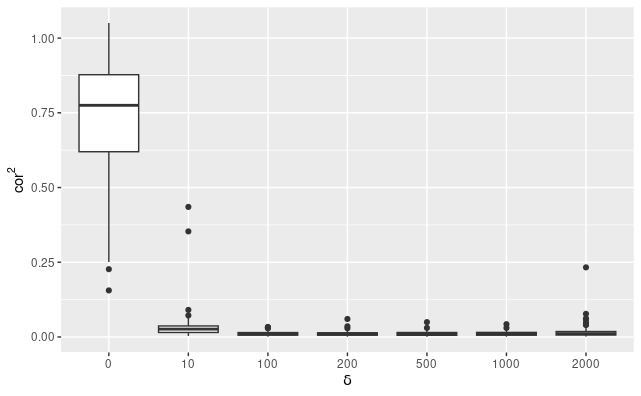}

  \includegraphics[width=0.5\textwidth]{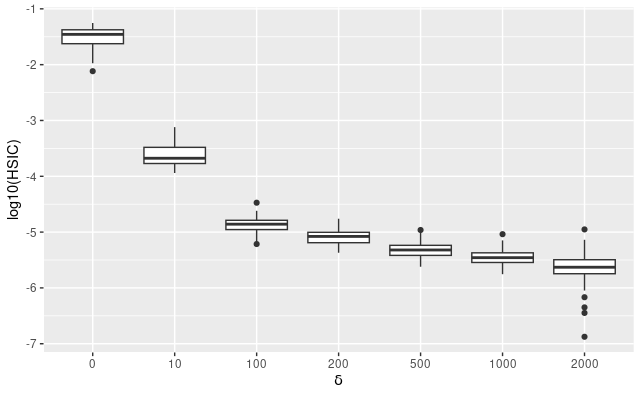}
  \includegraphics[width=0.5\textwidth]{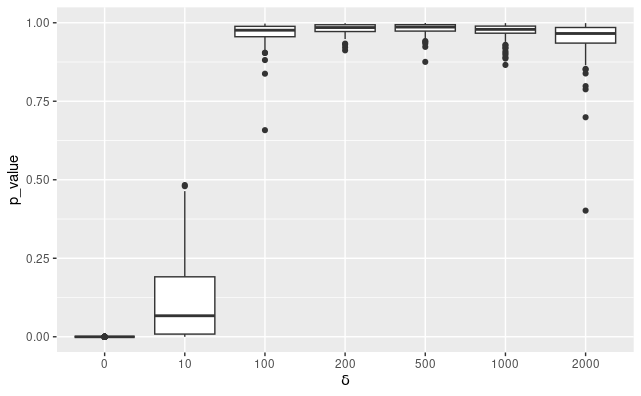}   

    \caption{Impact of the parameter $\delta$ on the $AUC$ (upper left), the norm of the correlation matrix (upper right), the log$_{10}$ HSIC (bottom left), and the p-value of the independence test (bottom right).}

\label{impact_delta_graph}
\end{figure}

Average and standard deviation of several metrics have been reported in \cref{table_delta}.  Increasing the $\delta$ parameters from 0 to 2000 decreases the $AUC$ in average from 0.758 to 0.708. However, the linear correlation and the HSIC between the latent space and the protected variable decreases to reach a value closer to 0. The more $\delta$ increases, the less the independence hypothesis is rejected. For $\delta=2000$, the algorithm is sometimes unstable. 

\begin{table}
\caption{Comparison of fair and bipartite variational graph auto-encoder for different value of $\delta$ on 100 trials with simulated data}
\centering
\resizebox{\textwidth}{!}{
\begin{tabular}{ |c|c|c|c|c|c|c|c| } 

 \hline
  $\delta$ & 0 & 10 & 100 & 200 & 500 & 1000 & 2000\\ 
 \hline
\rule{0pt}{2.5ex}AUC & $0.758\pm0.009$ & $0.735\pm0.012$ & $0.725\pm0.013$ & $0.724\pm0.012$ & $0.724\pm0.013$ & $0.723\pm0.013$ & $0.708\pm0.031$\\

\rule{0pt}{3ex} HSIC & $3.32 \times 10^{-2}$ & $2.75\times 10^{-4}  $ & $1.42 \times  10^{-5} $ & $8.63 \times  10^{-6} $ & $5.06 \times  10^{-6} $ & $3.69 \times  10^{-6} $  & $2.64 \times  10^{-6} $ \\ 

 & $\pm1.22 \times 10^{-2}$ & $\pm 1.56\times 10^{-4}  $ & $\pm4.37 \times  10^{-6} $ & $\pm2.72 \times  10^{-6} $ & $\pm1.65 \times  10^{-6} $ & $\pm1.27 \times  10^{-6} $  & $\pm1.50 \times  10^{-6} $ \\

$\rule{0pt}{3ex}\#p_{0.05}$ & 100/100  &44/100  &0/100  &0/100  & 0/100   & 0 /100 & 0/100 \\ 
$cor$ &$0.735\pm0.206$ & $0.035\pm0.055$ & $0.011\pm0.007$ & $0.011\pm0.008$ & $0.011\pm0.008$ & $0.011\pm0.008$ & $0.017\pm0.025$\\ 
 \hline
\end{tabular}}
\label{table_delta}
\end{table}

\subsection{Fair BGVAE with binary protected variable}
\label{binary}

The HSIC can encourage independence with respect to continuous variables or to categorical variables. The latter point is illustrated in this subsection.
\subsubsection{Setting}

Simulations with a similar setting as in \ref{SIMULATION} has been performed with a simulated latent space structured along a binary protected variable $S\in \{-1,1\}$.

In this simulation, we are going to generate a bipartite network made of $n_1=1000$ rows and $n_2 =100$ columns. 
Let $S_i \: {i.i.d.}$ for $i = 1,\dots,n_1$ with a Rademacher distribution ($\mathbb{P}(S_i = -1) = \mathbb{P}(S_i = 1) = \frac{1}{2} $) and $T_i \overset{i.i.d.}{\sim} \mathcal{N} (0,1)$ for $i = 1,\dots,n_1$ and independent of $S$. We suppose that $S$ is the protected variable. Let $Z_1 = (S,T) \in \mathbb{R}^{n_1 \times 2}$ be the 2-column matrix made with both $S$ and $T$. Let $Z_2\overset{i.i.d.}{\sim} \mathcal{N}\left(\begin{bmatrix} 0\\0 \end{bmatrix}, \begin{bmatrix} 1 & 0 \\ 0 & 1 \end{bmatrix}\right)\in \mathbb{R}^{n_2 \times2}$. We simulate our bipartite adjacency matrix with Bernoulli  $ B_{i,j} \overset{i.i.d.}{\sim}  \mathcal{B}(sigmoid(z_{1i}^\top\mathbf{I}_{D_+,D_-}z_{2j}))$.

First, we fit a classical bipartite and variational graph auto-encoder on $B_{i,j}$. We expect that this auto-encoder would yield a latent representation $\Tilde{Z_1}$ correlated with $S$ and $T$. 
We then fit our bipartite and fair auto-encoder to compare the result and see if the yielded latent space is independent of $S$.

\subsubsection{Results}

Results for the link prediction task in the simulated network are summarized in \cref{table_binary}.
The simulations were done with  dataset splits, with 30\% of the edges hidden. 20\% of these hidden edges are used as validation data set, and the remaining 10\% for the test set. Both sets also contain an equivalent amount of non-edges that are not in the train set.
In the table are reported the mean and standard deviation for 100 trials, except for $\#p_{0.05}$ which is only a count.
We set the hyperparameter $\delta = n_1=1000$.
For each trial, the simulations begin with 10 random initialization, and were fit using 1000 iterations of the Adam algorithm with learning rate $ 0.01$.  The model that achieved the most favorable performance on the validation test set is then selected to evaluate the performance on the test dataset.

\begin{figure}[h]
\centering
\includegraphics[width=0.5\textwidth]{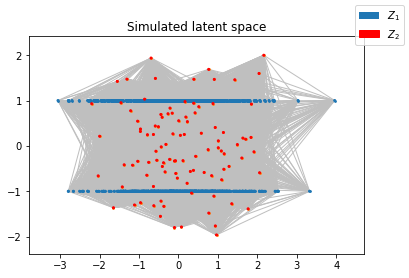}
\caption{Simulated latent space for generating bipartite network $B_{i,j}$. $Z_1 = (T,S)$ is represented in blue. $Z_2$ is represented in red and is independent of $Z_1$. The probability of connection between the node $i$ and $j$ will increase as the distance between their latent representation decreases. }
\label{fig_binary1}
\end{figure}

\begin{figure}[h]
\centering
\includegraphics[width=0.45\textwidth]{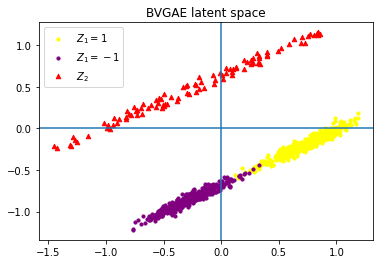}
\includegraphics[width=0.45\textwidth]{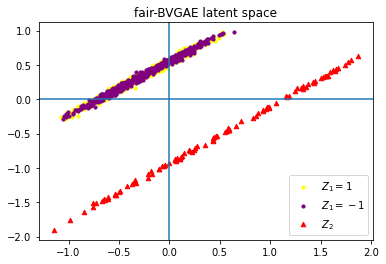}

\caption{Estimated latent space for the bipartite variational graph auto-encoder (left) and the fair bipartite variational graph auto-encoder (right) in the binary case.}
\label{figdiscretevar}
\end{figure}

As shown in Figure \ref{figdiscretevar},
we are able to provide an embedding independent of the binary variable with our fair VGAE contrary to the embedding provided by the simple VGAE. Average values and standard deviation of several metrics are reported in \cref{table_binary}. As expected, the average AUC decreases in the fair model compared to the classical case, however in the fair case, we do not reject the hypothesis of independence between the latent space and the protected variable. The adversarial setting decreases the correlation and the HSIC compared to the classical BVGAE, but the independence hypothesis has been rejected 46\% of the time.

\begin{table}
\caption{\label{table_binary}Comparison between the Bipartite variational graph auto-encoder and its fair counterparts on 100 trials with binary protected variable}
\resizebox{\textwidth}{!}{%
\begin{tabular}{ |c|c|c|c| } 

 \hline
   & BVGAE & fair-BVGAE & ADV \\ 
 \hline
AUC & $0.774 \pm 0.012$ & $0.634 \pm 0.061$ & $0.634 \pm 0.073$ \\ 
HSIC & $9.32\times 10^{-2}\pm 0.42\times 10^{-2} $  & $1.37\times 10^{-6}\pm 1.50\times 10^{-6} $ & $5.17\times 10^{-3}\pm 13.2\times 10^{-3} $\\ 
$\#p_{0.05}$ & 100/100     & 0/100  & 54/100 \\ 
$cor$ & $0.964 \pm 0.020 $  & $0.014\pm 0.029$  & $0.246 \pm 0.322 $ \\ 
 \hline
\end{tabular}
}
\end{table}

\subsection{Temporal gain using RFF HSIC}
\label{HSIC_temporal_gain}
The purpose of this study is to see the temporal gain of using the RFF HSIC instead of $\widehat{HSIC}$ for different value of $n$, and to see if $RFF\ HSIC$ is an accurate approximation of  $\widehat{HSIC}$.

\subsubsection{Settings}
For various value of $n$, we consider $X$ a $n\times 4$ matrix, with entries such as $X_{i,j} \overset{i.i.d.}{\sim} \mathcal{N}(0,1)$. Let $S$ be a $n \times 4$ matrix. Under the null hypothesis, we consider that $S$ is independent of $X$, with $S_{i,j} \overset{i.i.d.}{\sim} \mathcal{N}(0,1)$. Under the alternative hypothesis we consider that $S= 3X$. The aim is to compute the HSIC between $X$ and $S$ with $\widehat{HSIC}(X,S)$, $RFF\ HSIC(X,S)$ and their respective gradient, under both hypothesis.
In our fairness setting, $S$ would represent the protected variable and would be fixed once and for all, while $X$ would change according to the computed gradient. Therefore, we evaluate $L_{i,j} = L(s_i,s_j) = e^{-\frac{||s_i-s_j||^2}{2}}$, $L' = \sum_{1\leq p,q\leq n} L_{q,p}$, and $L''_{i} = \sum_{q =1}^n L_{i,q}$ in advance to perform a quicker computation :  

\begin{align}
&\widehat{HSIC}=  \notag\\
&=\frac{1}{n^2}\sum_{1\leq i,j\leq n}K_{i,j}L_{i,j}
+\frac{1}{n^4}\sum_{1\leq i,j\leq n}K_{i,j} \underbrace{\sum_{1\leq p,q\leq n} L_{p,q}}_{L'} 
- \frac{2}{n^3}\sum_{1\leq i,j\leq n}K_{i,j}\underbrace{\sum_{q=1}^n L_{i,q}}_{L''i}\\
&=\frac{1}{n^2}\sum_{1\leq i,j\leq n}K_{i,j}L_{i,j}+ 
\frac{L'}{n^4}\sum_{1\leq i,j\leq n}K_{i,j}
- \frac{2}{n^3}\sum_{1\leq i,j\leq n}K_{i,j}L''_i.
\label{eq_HSIC}
\end{align}
For multiple random realisations of $X$, we compute $K_{i,j} = e^{-\frac{||x_i-x_j||^2}{2}}$ and then plug it in \cref{eq_HSIC}. We also compute the gradient with respect to $X$ using Pytorch automatic differentiation, before measuring the average time taken to realize both of these operations.

Using the method presented in \cref{eq_RFF}, we also compute the $RFF\ HSIC$ between $X$ and $S$ with $h = \left\lceil\sqrt{n}\right\rceil$, and its gradient using Pytorch automatic differentiation. We measure the average time taken to calculate the $RFF\ HSIC$ and its gradient compared to the $\widehat{HSIC}$. We also compare how close the value of $RFF\ HSIC$ is to the $\widehat{HSIC}$ using the squared error between the two with $\widehat{HSIC}$ as a base value. All theses computation are realized on an Intel Xeon(R) CPU E5-1650 v4 and 32GB of RAM.

\subsubsection{Results}
\begin{figure}
    \centering
    \includegraphics[width= 0.95\textwidth]{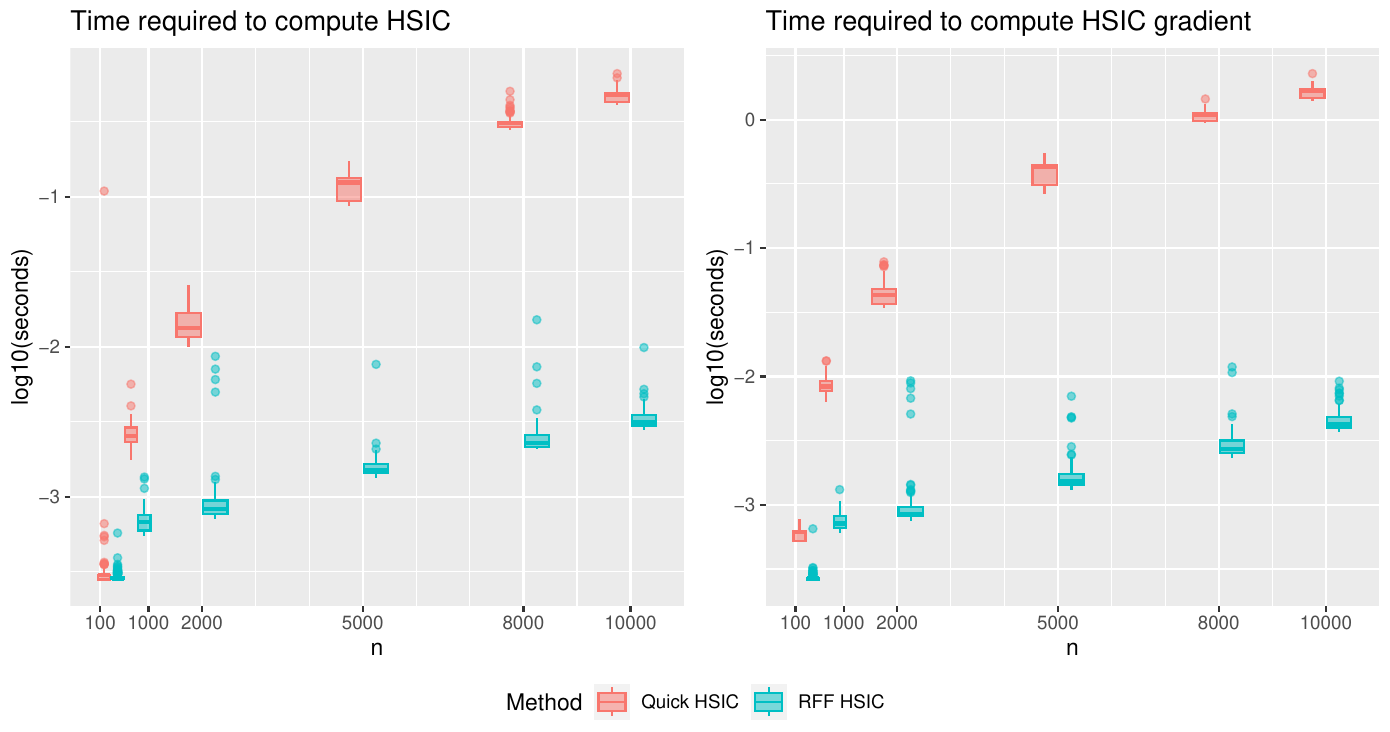}
   
    \caption{Estimated time to compute $HSIC(X,S)$ (left) and its gradient with respect to $X$(right) using $\widehat{HSIC}$ and $RFF\ HSIC$.}
    \label{HSIC_timer}
\end{figure}

\begin{figure}
    \centering
    \includegraphics[width= 0.75\textwidth]{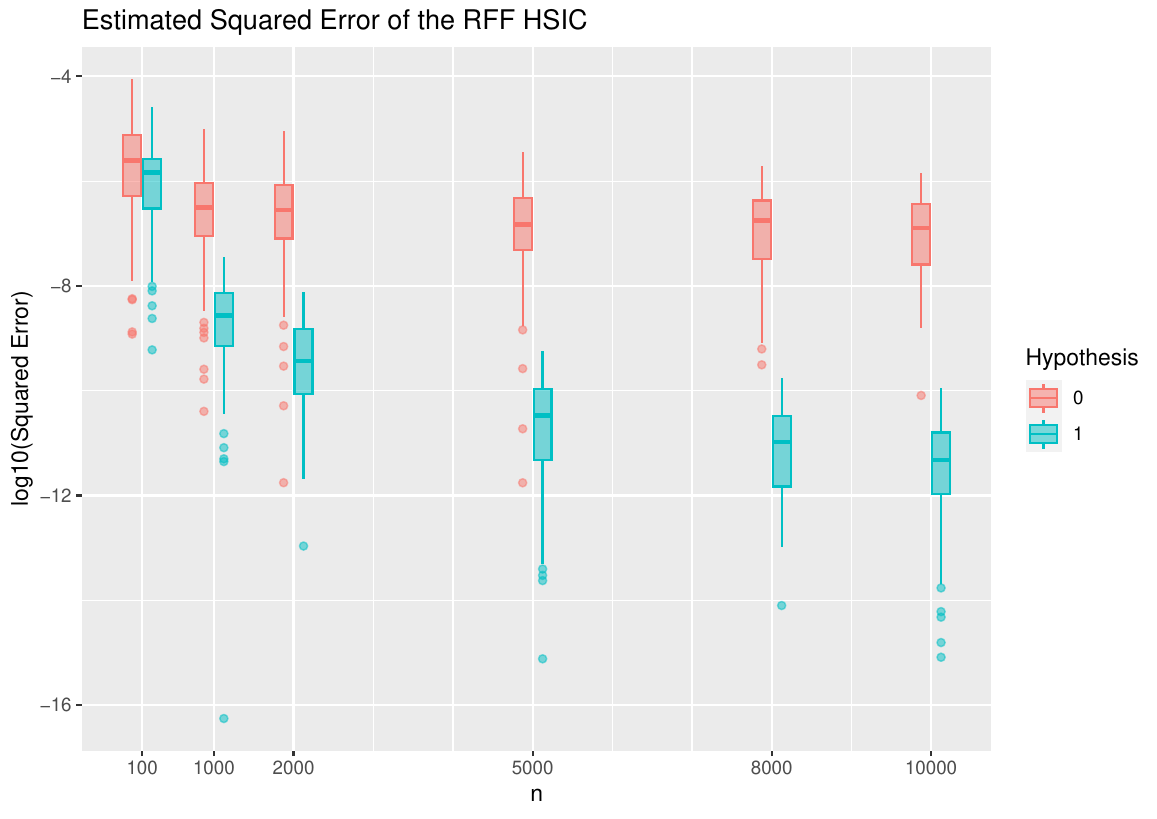}
    \caption{Estimated squared error between the $RFF\ HSIC$ and the $\widehat{HSIC}$}
    \label{HSIC_acc}
\end{figure}

As we can see in \cref{HSIC_timer}, using the $RFF\ HSIC$ is much faster than $\widehat{HSIC}$ by a large margin. Under the null hypothesis, the estimation is less accurate than under the alternative hypothesis (\cref{HSIC_acc}) but the hypothesis doesn't affect the computation time. For 1000 iterations and for $n=10000$, the $RFF\ HSIC$ and its gradient would require around 7.9 seconds of computation time, while the $\widehat{HSIC}$ would require around 35 minutes. In the Spipoll dataset, we have considered $n=12574$ with a latent space of dimension 4. Using a second order polynomial, we can estimate that computing $1000$ times the HSIC and its gradient would require around 56 minutes for $\widehat{HSIC}$ and 11 seconds for $RFF\ HSIC$. We also only presented results from data in the time period between 2017 and 2020, but if we considered the Spipoll data set from 2010 to 2020, then $n \approx 26000$. In this case, we can estimate that computing $1000$ times the HSIC and its gradient would require around 4 hours with $\widehat{HSIC}$ while the $RFF\ HSIC$ would only need 33 seconds. All these estimations are done without taking into account the fact that the computation of the $n \times n$ Gram matrix, needed for the $\widehat{HSIC}$ can also require a lot of memory from the computer.

\clearpage
\section{Latent space representation of Spipoll}
\label{original}

\begin{figure}[h]
    \centering
    \includegraphics[width = 0.8\textwidth]{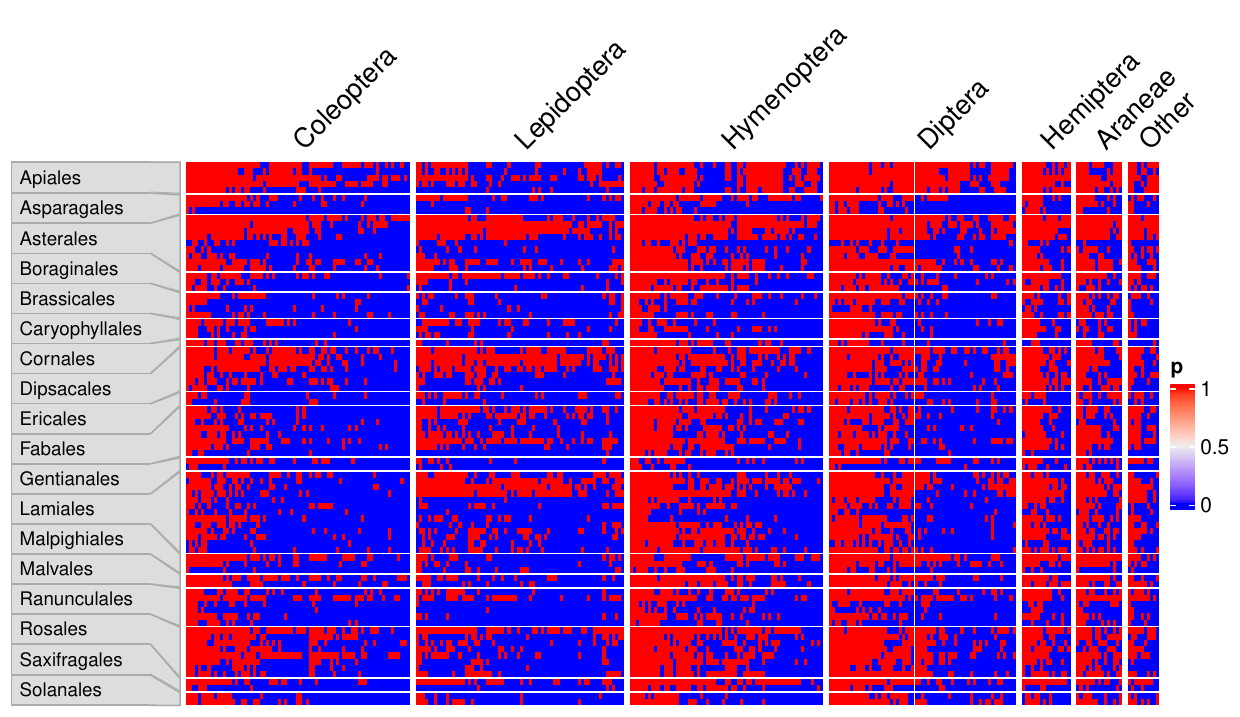}
    \caption{Observed initial plant-pollinator network}
    \label{observed_initial}
\end{figure}
The observed plant-pollinator network is provided in \cref{observed_initial}.
In addition to the reconstructed plant-pollinator network, the
method provides an embedding of dimension $D = D_+ + D_- = 4$ with $D_+ = D_- = 2$, which means that for the first two dimensions, insects and sessions that are embedded in the same direction are more likely to be connected, and the ones in the opposite direction are less likely to be connected. On the contrary, insects and sessions that are embedded in the same direction for the third and fourth dimensions are less likely to be connected, while the ones in the opposite direction are more likely to be connected. The choice of $D_+ = D_- = 2$ is justified by looking at \cref{Z_dim}. The session-pollinator embedding can be seen in \cref{spipoll1} and \cref{spipoll2}.

\clearpage

\subsection{Spipoll, exploration with higher dimensional latent spaces}
\label{spipoll_appendix}
\begin{figure}[h]
    \centering
    \includegraphics[width = 0.4\textwidth]{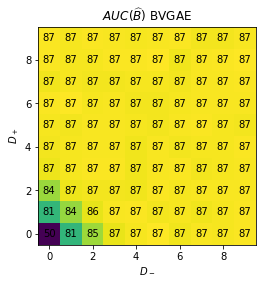}
    \includegraphics[width = 0.4\textwidth]{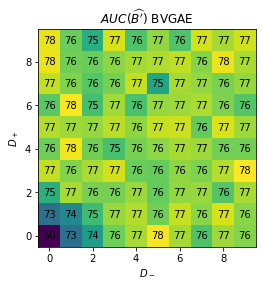}
    \includegraphics[width = 0.4\textwidth]{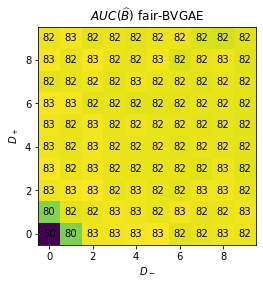}
    \includegraphics[width = 0.4\textwidth]{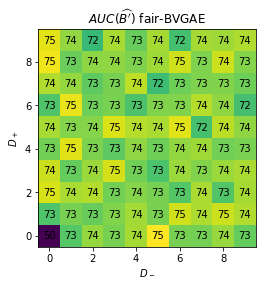}
    \caption{Estimated mean on 10 trials for the $AUC(\widehat{B})$ (left) and $AUC(\widehat{B'})$ (right) for link prediction in the Spipoll data set using BGVAE (top) and the fair-BGVAE (bottom) for various values of $D_+$ and $D_-$.}
    \label{Z_dim}
\end{figure}
In \cref{spipoll_result_section} we show in detail the results for the case where the latent space has 4 dimensions with $D_+ = D_- = 2$. We justify this choice by looking at the estimated mean of the $AUC(\widehat{B})$ for different numbers of dimensions for the latent space. Looking at \cref{Z_dim}, we can see that the $AUC(\widehat{B})$  doesn't significantly change for higher values of $D_+$ and $D_-$.

\clearpage

\subsection{Latent space representation}

\begin{figure}[h]
    \centering
    \includegraphics[width = \textwidth]{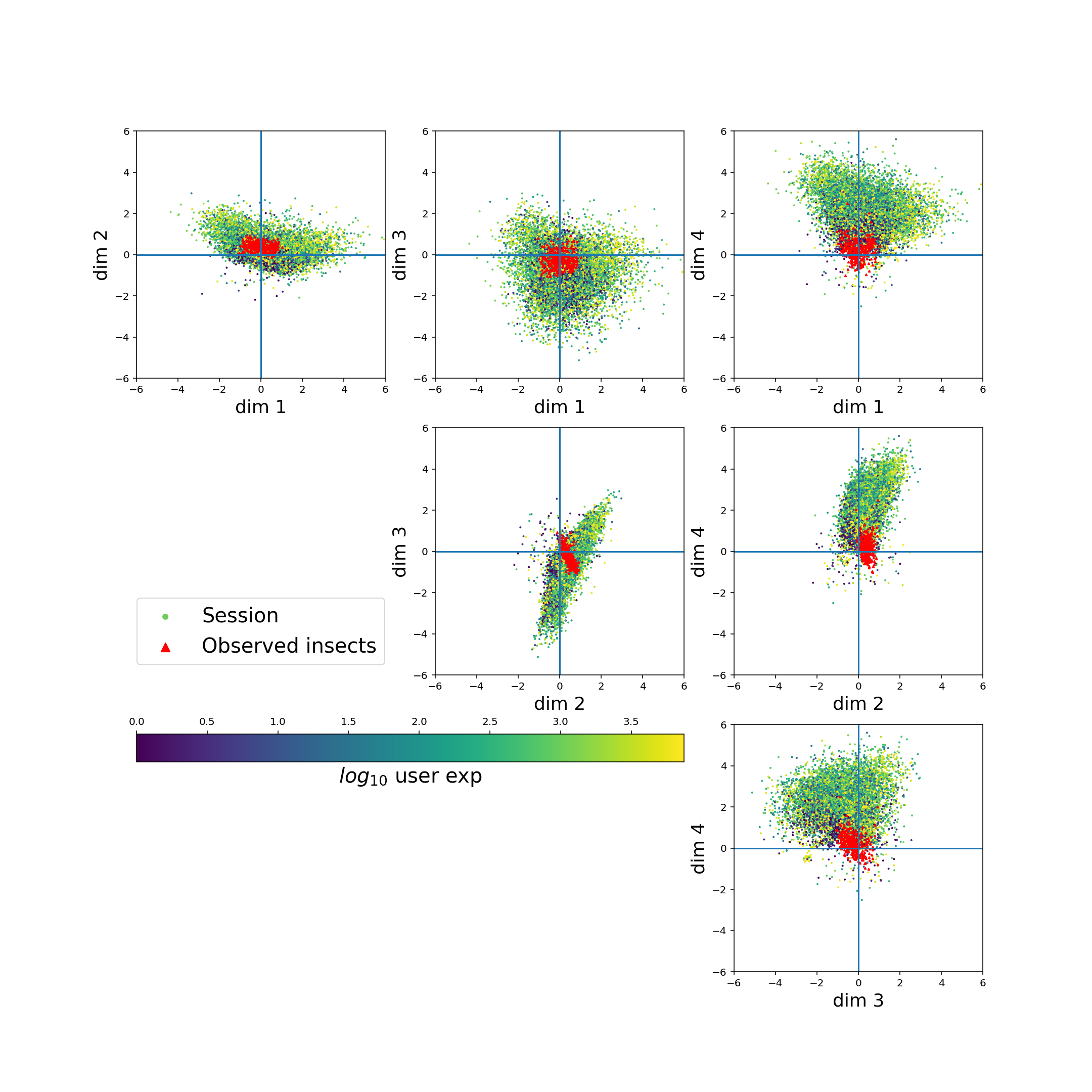}
\caption{Estimated latent space for the Spipoll data set using BVGAE}
\label{spipoll1}
\end{figure}

\clearpage
    \begin{figure}[h]
        \centering
    \includegraphics[width = \textwidth]{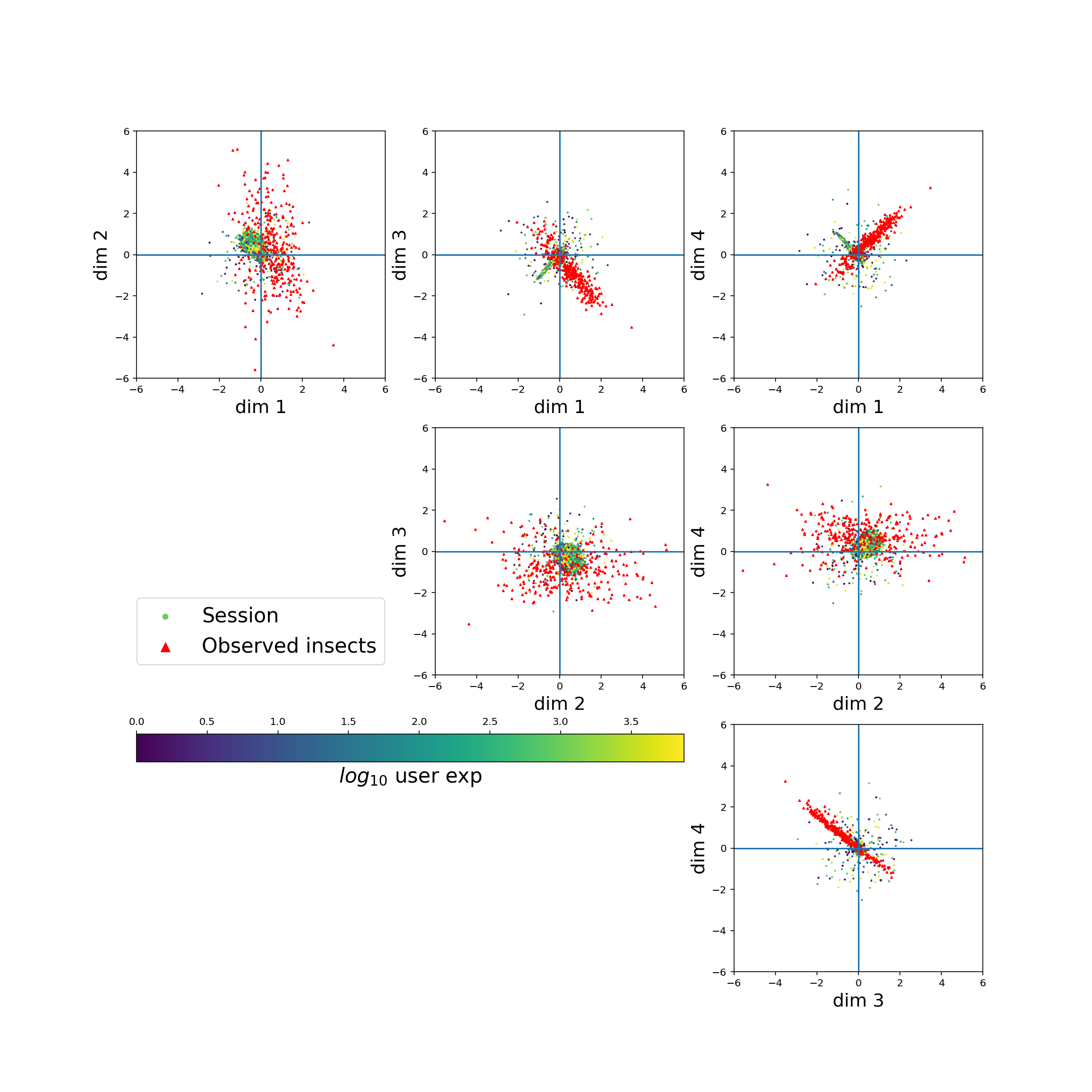}
\caption{Estimated latent space for the Spipoll data set using fair-BVGAE}
\label{spipoll2}
\end{figure}

\end{document}